**PROJECT**

**ON**

# A POLYNOMIAL BASED DEPTH ESTIMATION FROM A SINGLE IMAGE

**By**

**Kaushik K. Tiwari**

## TABLE OF CONTENTS

## ABSTRACT

The depth of a visible surface of a scene is the distance between the surface and the sensor. Recovering depth information from two-dimensional images of a scene is an important task in computer vision that can assist numerous applications such as object recognition, scene interpretation, obstacle avoidance, inspection and assembly. Various passive depth computation techniques have been developed for computer vision applications. They can be classified into two groups. The first group operates using just one image. The second group requires more than one image which can be acquired using either multiple cameras or a camera whose parameters and positioning can be changed.

The process is aimed to find the real depth of the object from the camera which had been used to click the photograph. An n-degree polynomial was formulated, which maps the pixel depth of an image to the real depth. In order to find the coefficients of the polynomial, an experiment was carried out for a particular lens and thus, these coefficients are a unique feature of a particular camera.

## LITERATURE SURVEY

Depth finding by camera and image processing have variant applications, including industry, robots and vehicles navigation and controlling. This issue has been examined from different viewpoints, and a number of researches have conducted some valuable studies in this field. All of the introduced methods can be categorized into six main classes.

The first class includes all methods that are based on using two cameras. These methods origin from the earliest researches in this field that employ the characteristics of human eye functions. In these methods, two separate cameras are stated on a horizontal line with a specified distance from each other and are focused on a particular object. Then the angles between cameras and the horizontal line are measured, and by using triangulation methods, the vertical distance of the object from the line connecting two cameras is calculated. The main difficulty of these methods is the need to have mechanical moving and the adjustment of the cameras in order to provide proper focusing on the object. Another drawback is the need of the two cameras, which will bring more cost and the system will fail if one of them fails.

The second class emphasize on using a single camera . In these methods, the base of the measurement is the amount of the image resizing in proportion to the camera movement. These methods need to know the main size of the object subjected to distance measurement and the camera's parameters such as the focal length of its lens.

The methods in the third class are used for measuring the distance of the moving targets . In these methods, a camera is mounted on a fixed station. Then the moving object(s) is(are) indicated, based on the four scenarios: maximum velocity, small velocity changes, coherent motion, continuous motion. Finally, the distance of the specified target is calculated. The major problem in these methods is the large amount of the necessary calculations.

The fourth class includes the methods which use a sequence of images captured with a single camera for depth perception based on the geometrical model of the object and the camera . In these methods, the results will be approximated. In addition, using these methods for the near field (for the objects near to the camera) is impossible.

The fifth class of algorithms prefer depth finding by using blurred edges in the image . In these cases, the basic framework is as follows: The observed image of an object is modeled as a result of convolving the focused image of the object with a point spread function. This point spread function depends both on the camera parameters and the distance of the object from the camera. The point spread function is considered to be rotationally symmetric (isotropic). The line spread function corresponding to this point spread function is computed from a blurred step-edge. The measure of the spread of the line spread function is estimated from its second central moment. This spread is shown to be related linearly to the inverse of the distance. The constants of this linear relation are determined through a single camera calibration procedure. Having computed the spread, the distance of the object is determined from the linear relation.

In the last class, auxiliary devices are used for depth perception. One of such methods uses a laser pointer which three LEDs are placed on its optical axis], built in a pen-like device. When a user scans the laser beam over the surface of the object , the camera captures the image of the three spots (one for from the laser, and the others from LEDs), and then the triangulation is carried out using the camera's viewing direction and the optical axis of the laser. The main problem of these methods is the need for the auxiliary devices, in addition to the camera, and consequently the raise of the complexity and the cost.

## 1.1 Monocular Cues

There are also numerous monocular cues—such as texture variations and gradients, defocus, color/haze, etc.—that contain useful and important depth information. Estimating depth from a single image using monocular cues requires a significant amount of prior knowledge, since there is an intrinsic ambiguity between local image features and depth variations. Depth estimation from monocular cues is a difficult task, which requires that we take into account the global structure of the image. applied supervised learning to the problem of estimating depth from single monocular images of unconstrained outdoor and indoor environments. The image is divided into small rectangular patches, and estimate a single depth value for each patch.Two types of features are used : absolute features—used to estimate the absolute depth at a particular patch—and relative features, which we use to estimate relative depths (magnitude of the difference in depth between two patches).

## 1.2 Stereo Cues

Depth Estimation in computer vision and robotics is most commonly done via stereo vision (stereopsis), in which images from two cameras are used to triangulate and estimate distances. The two images are taken from a pair of stereo cameras. When the scene is imaged by two cameras at two different positions, their images will be disparate.

The stereopsis method is based on measuring this disparity. Then, the triangulation method is used to recover the 3-D structure. Over the past few decades, researchers have developed very good stereo vision systems. Although these systems work well in many environments, stereo vision is fundamentally limited by the baseline distance between the two cameras. Specifically, the depth estimates tend to be inaccurate when the distances considered are large (because even very small triangulation/ angle estimation errors translate to very large errors in distances). Further, stereo vision also tends to fail for texture less regions of images where correspondences cannot be reliably found. Stereo cues are based on the difference between two images and do not depend on the content of the image. The images can be entirely random, and it will generate a pattern of disparities

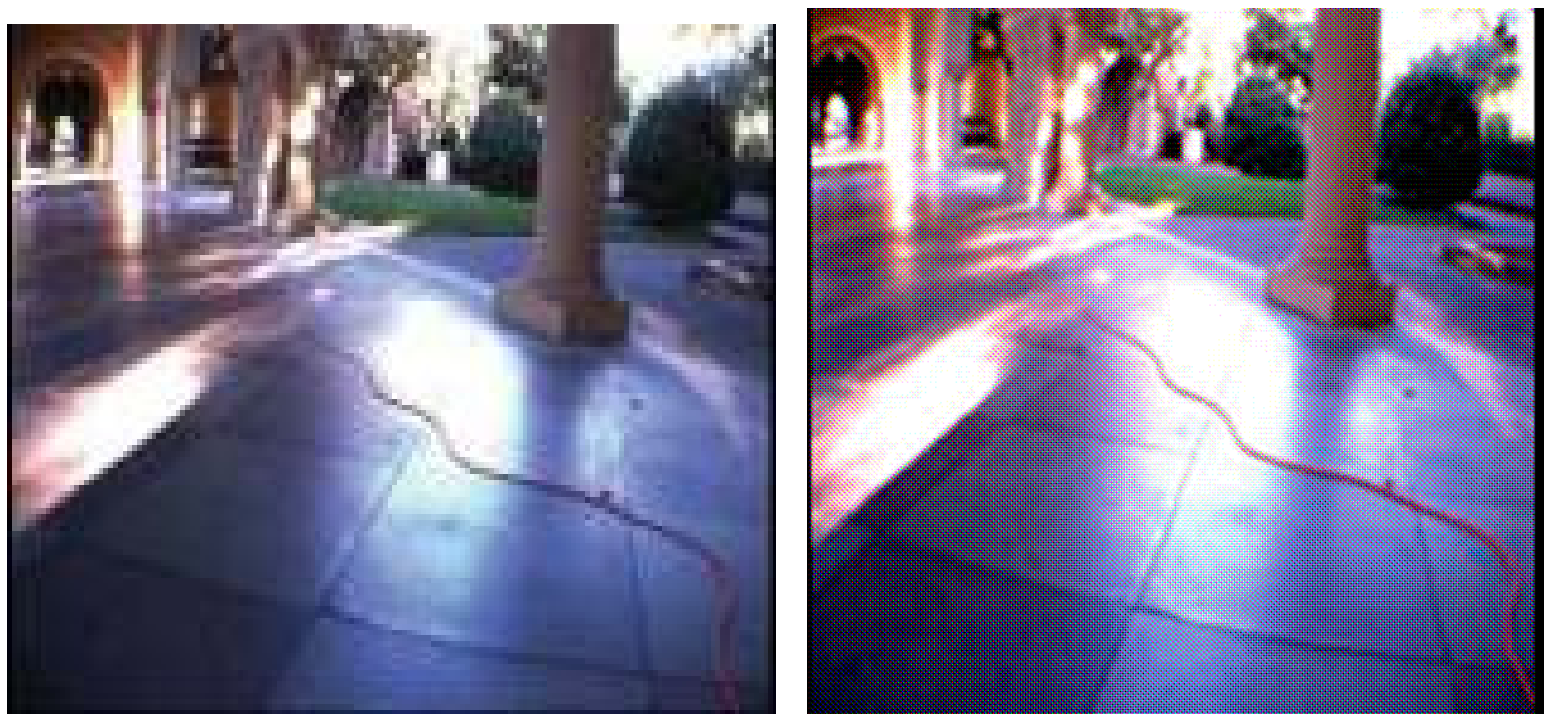

Figure 1: Two images of the same scene taken from a pair of stereo cameras

## 1.3 Camera Focus

When a point object is photographed from a camera , the image produced is also a point object if the camera is correctly focused at by the photographer. In case the camera is not properly focused, the image produced of the point object is a circular disc constituting the blurred region with some thickness. Figure 2 illustrates that the width *b* of blurred image of a point object P is  a function of model parameters *d*, *w*, *c*, lens one *U*, and object distance *s*.

From the geometric relation of these parameters, *b* is obtained by

$$b = wd\left|\frac{1}{s+c} - \frac{1}{U+c}\right| = \frac{w^2}{B}\left|\frac{1}{s+c} - \frac{1}{U+c}\right|$$

where $B = w/d$.

Using this equation, we have the depth *s* of an object

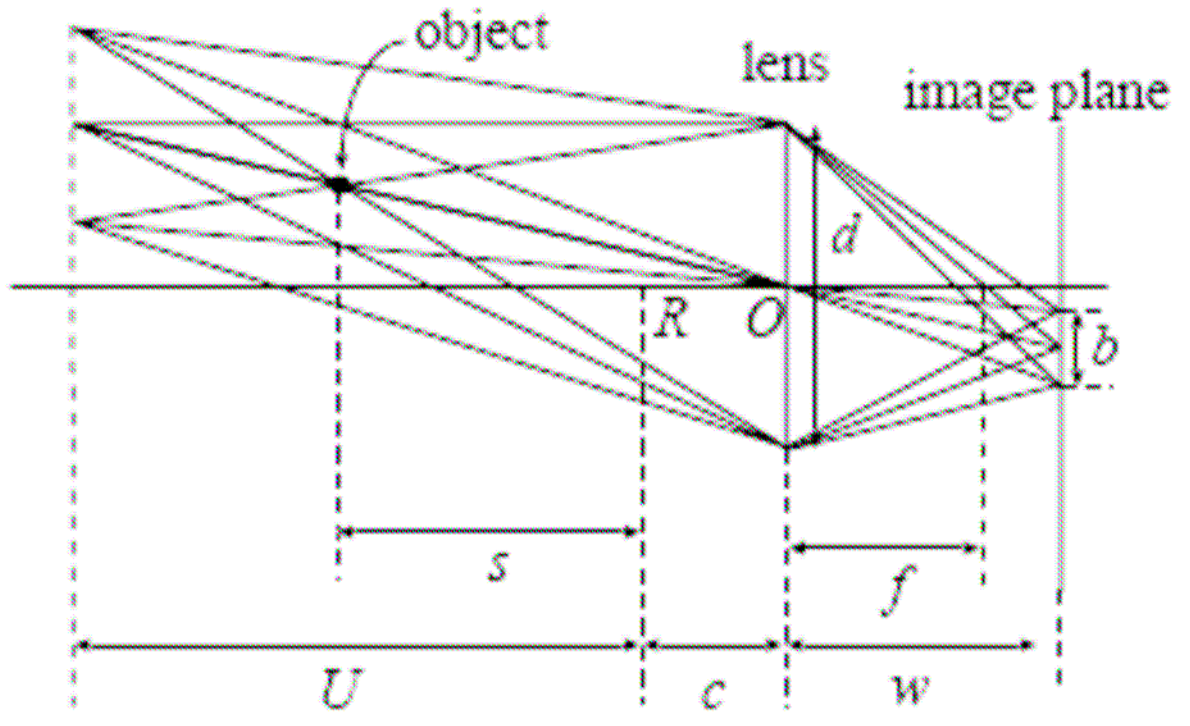

Figure 2: Width of blur *b*  of the point object

from its width of blur *b*, that is,

$$s = \begin{cases} \dfrac{wd(U+c)}{wd+b(U+c)} - c & (s < U) \\ \dfrac{wd(U+c)}{wd-b(U+c)} - c & (s > U) \end{cases}$$

where *w*, *c*, *d* and *U* are given by zoom, focus and iris settings. Thus, the depth *s* is computed from the width of blur *b* at any setting of the three real parameters. From a set of multiple images taken at different parameters, we can have reliable depth information.

## 1.4 Depth from Lens Translation

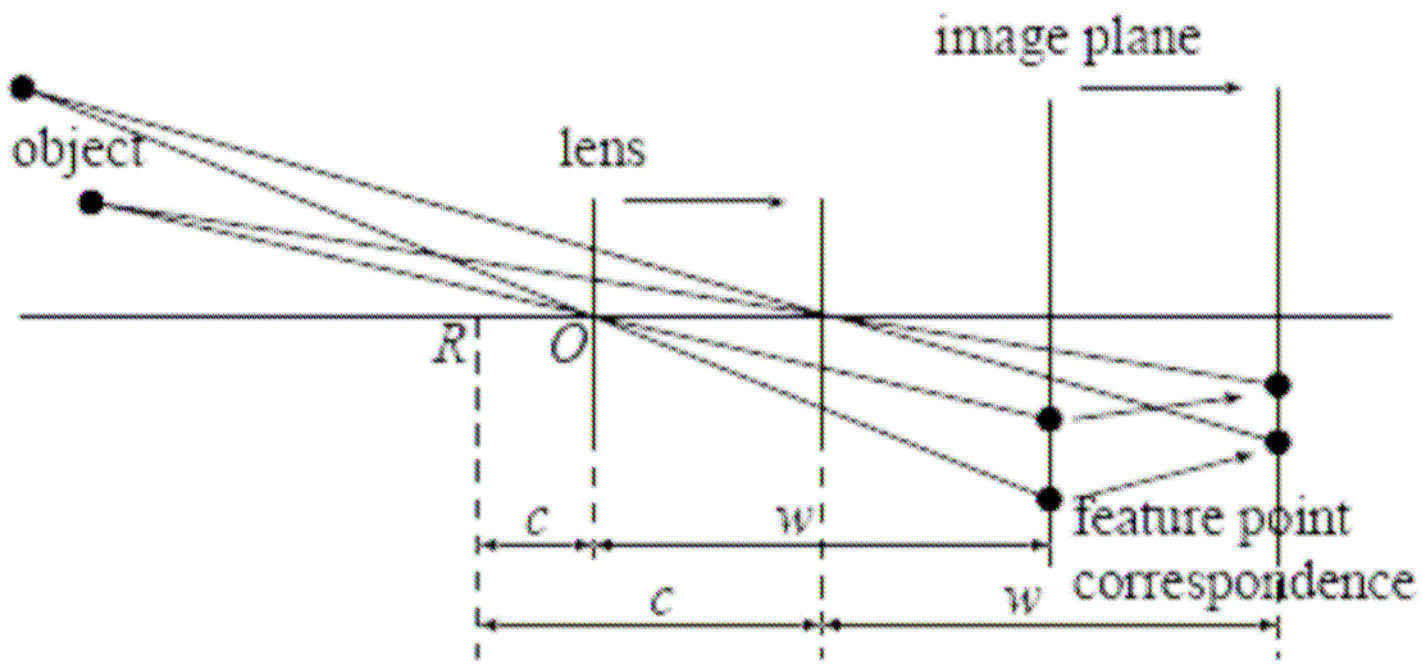

Figure 3 : Depth from Lens Translation

Depth from lens translation is a particular case of "stereo" or "shape from motion" problem; that is, the depth information is derived from the correspondence of feature points detected in images. Figure 3 illustrates the lens center translation along the optical axis of the lens and the correspondence of the projected object points between two images. In general, the $p^{th}$ object point $\boldsymbol{sp}$ = $(Xp, Yp, Zp)T$ represented in the world coordinates is perspectively projected on the $f^{th}$ image coordinates as $\boldsymbol{ufp}$ = $(xfp, yfp)T$ .

$$u_{fp} = \mathcal{P}(M_f s_p + t_f)$$

where $Mf$ and $\boldsymbol{tf}$ denote the rotation matrix and translation vector respectively, and $P$ is defined as a perspective projection operator with effective focal length $w$

$$\mathcal{P}\begin{bmatrix} x \\ y \\ z \end{bmatrix} = \frac{w}{z}\begin{bmatrix} x \\ y \end{bmatrix}.$$

Depth recovery, or 3D shape reconstruction in other words, is performed by using the least squares criterion by minimizing the following error function.

$$Error = \sum_{f,p} ||\tilde{u}_{fp} - u_{fp}||^2$$

where $\tilde{\boldsymbol{u}}_{fp}$ represents the $p^{th}$ feature point detected in the $f^{th}$ image and $\boldsymbol{u}_{fp}$ is the feature point computed from the estimated 3D object point. In order to reduce the computational complexity, we convert the images to be taken by a camera that has a fixed *w*. This means that the image size is first modified then the depth estimation is performed.

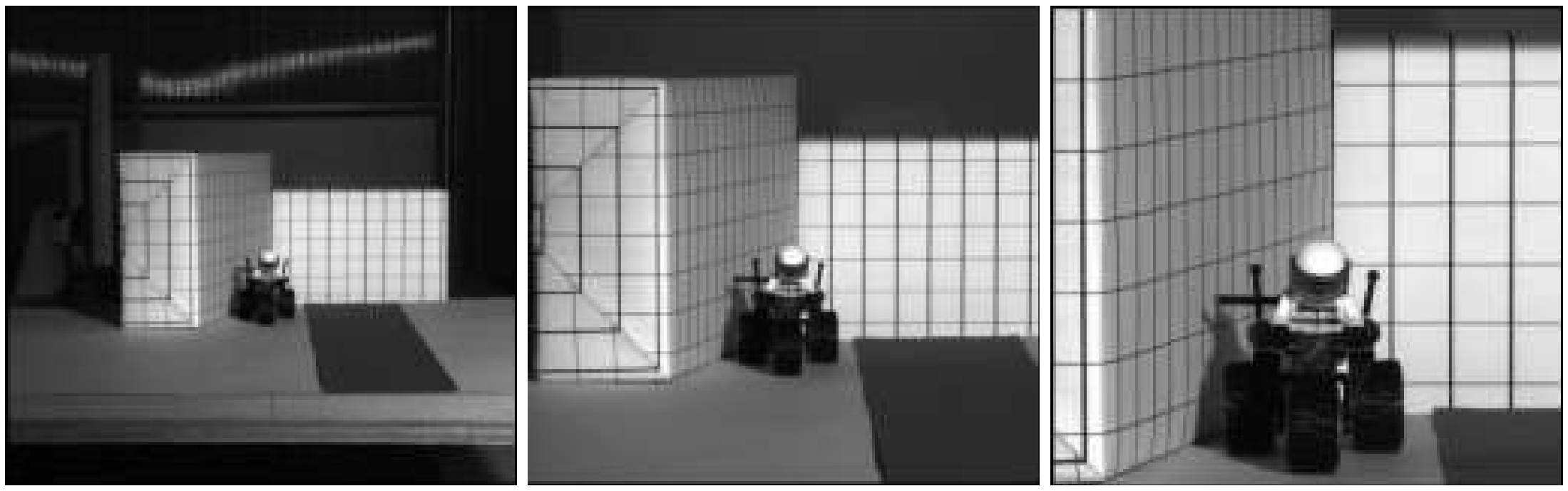

Images with lens translation

**1.5 Size of Objects**

The literature on absolute depth estimation is very large but the proposed methods rely on a limited number of sources of information, such as, binocular vision, motion parallax, defocus etc. However, under normal viewing conditions, observers can provide a rough estimate of the absolute depth of the scene even in the absence of all these sources of information ( eg. While looking at a photograph ). One additional source of information for absolute depth estimation is the use of size of familiar objects like faces, bodies, cars etc. however, this strategy requires decomposing the image into its constituent elements. The process of image segmentation and object recognition, under unconstrained conditions, still remains difficult and unreliable for current computational approaches.

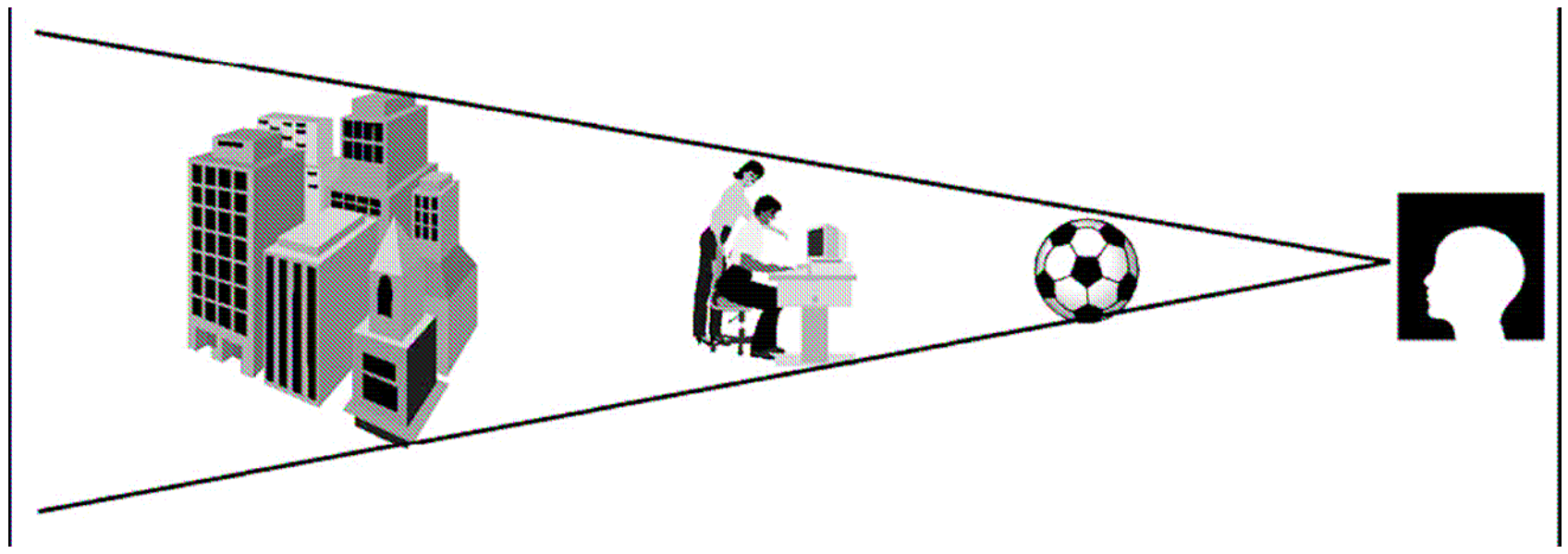

Figure 4 :The recognition of image structure provides unambiguous monocular information about the depth between the observer and the scene.

**Image processing** is any form of information processing for which both the input and output are images, such as photographs or frames of video. Most image processing techniques involve treating the image as a two-dimensional signal and applying standard signal processing techniques to it. Work on image can be divided into three categories:

- Image Processing *image in -> image out*
- Image Analysis *image in -> measurements out*
- Image Understanding *image in -> high-level description out*

Earlier image processing was done largely in the analog domain, chiefly by optical devices. These optical methods are still essential to applications such as holography because they are inherently parallel; however, due to the significant increase in computer speed, these techniques are increasingly being replaced by digital image processing methods.

Digital image processing techniques are generally more versatile, reliable, and accurate; they have the additional benefit of being easier to implement than their analog counterparts. Specialized hardware is still used for digital image processing: computer architectures based on pipelining have been the most commercially successful. There are also many massively parallel architectures that have been developed for the purpose. Today, hardware solutions are commonly used in video processing systems. However, commercial image processing tasks are more commonly done by software running on conventional personal computers.

## 2.1 Introduction To Images:

An image defined in the "real world" is considered to be a function of two real variables, for example, $a(x,y)$ with $a$ as the amplitude (e.g. brightness) of the image at the *real* coordinate position $(x,y)$. An image may be considered to contain sub-images sometimes referred to as *regions–of–interest*, *ROIs*, or simply *regions*. This concept reflects the fact that images frequently contain collections of objects each of which can be the basis for a region. In a sophisticated image processing system it should be possible to apply specific image processing operations to selected regions. Thus one part of an image (region) might be processed to suppress motion blur while another part might be processed to improve color rendition.

The amplitudes of a given image will almost always be either real numbers or integer numbers. The latter is usually a result of a quantization process that converts a continuous range (say, between 0 and 100%) to a discrete number of levels. In certain image-forming processes, however, the signal may involve photon counting which implies that the amplitude would be inherently quantized. In other image forming procedures, such as magnetic resonance imaging, the direct physical measurement yields a complex number in the form of a real magnitude and a real phase.

## 2.2 Digital Image Processing:

A digital image $a[m,n]$ described in a 2D discrete space is derived from an analog image $a(x,y)$ in a 2D continuous space through a *sampling* process that is frequently referred to as digitization. The effect of digitization is shown in Figure 5.

The 2D continuous image $a(x,y)$ is divided into $N$ *rows* and $M$ *columns*. The intersection of a row and a column is termed a *pixel*. The value assigned to the integer coordinates $[m,n]$ with $\{m=0,1,2,...,M-1\}$ and $\{ n=0,1,2,...,N-1 \}$ is $a[m,n]$. In fact, in most cases $a(x,y)$--which we might consider to be the physical signal that impinges on the face of a 2D sensor--is actually a function of many variables including depth ($z$), color ($\lambda$), and time ($t$).

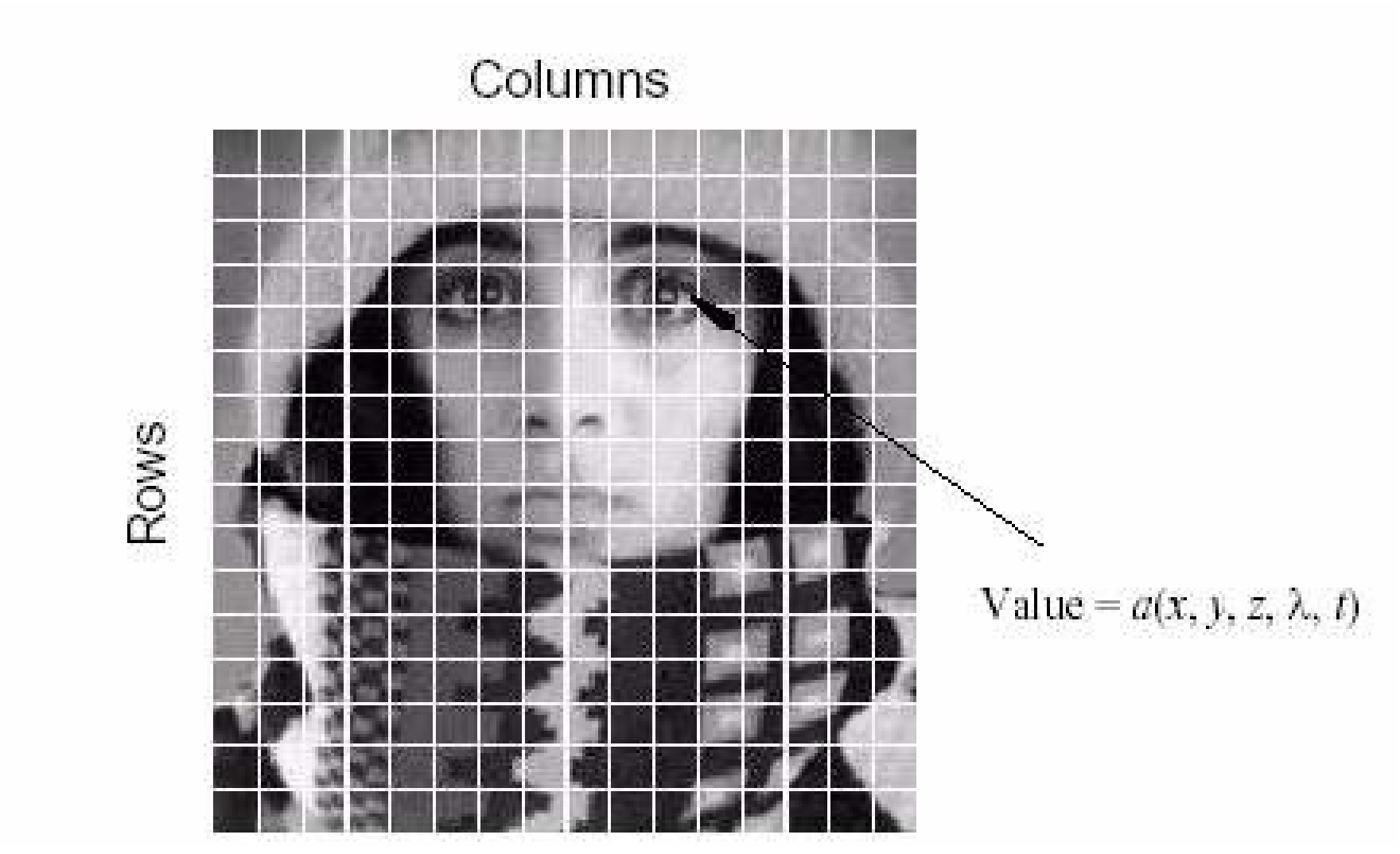

Figure 5 : Digitization of a continuous image. The pixel at coordinates [$m$=10, $n$=3] has the integer brightness value 110.

The image shown in Figure 5 has been divided into $N$ = 16 rows and $M$ = 16 columns. The value assigned to every pixel is the average brightness in the pixel rounded to the nearest integer value. The process of representing the amplitude of the 2D signal at a given coordinate as an integer value with $L$ different gray levels is usually referred to as amplitude quantization or simply *quantization*.

### 2.2.1 Common Values:

There are standard values for the various parameters encountered in digital image processing. These values can be caused by video standards, by algorithmic requirements, or by the desire to keep digital circuitry simple. Table 2.1 gives some commonly encountered values.

| *Parameter* | *Symbol* | *Typical values* |
|---|---|---|
| Rows | $N$ | 256,512,525,625,1024,1035 |
| Columns | $M$ | 256,512,768,1024,1320 |
| Gray Levels | $L$ | 2,64,256,1024,4096,16384 |

Table 2.1 Various Values Of the RGB,Gray levels

Quite frequently the value of $M=N=2^K$ where $\{K = 8,9,10\}$. This can be motivated by digital circuitry or by the use of certain algorithms such as the (fast) Fourier transform. The number of distinct gray levels is usually a power of 2, that is, $L=2^B$ where $B$ is the number of bits in the binary representation of the brightness levels. When $B>1$ we speak of a *gray-level image*; when $B=1$ we speak of a *binary image*. In a binary image there are just two gray levels which can be referred to, for example, as “black” and “white” or “0” and “1”.

### 2.2.2 Characteristics Of Image Operations

#### 2.2.2.1 Types of operations:

The types of operations that can be applied to digital images to transform an input image $a[m,n]$ into an output image $b[m,n]$ (or another representation) can be classified into three categories as shown in Table 2.2

| Operation | Characterization | Generic Complexity/Pixel |
|---|---|---|
| • *Point* | – the output value at a specific coordinate is dependent only on the input value at that same coordinate. | *constant* |
| • *Local* | – the output value at a specific coordinate is dependent on the input values in the *neighborhood* of that same coordinate. | $P^2$ |
| • *Global* | – the output value at a specific coordinate is dependent on all the values in the input image. | $N^2$ |

Table 2.2: Types of image operation. Image size = N x N

This is shown graphically in Figure 6.

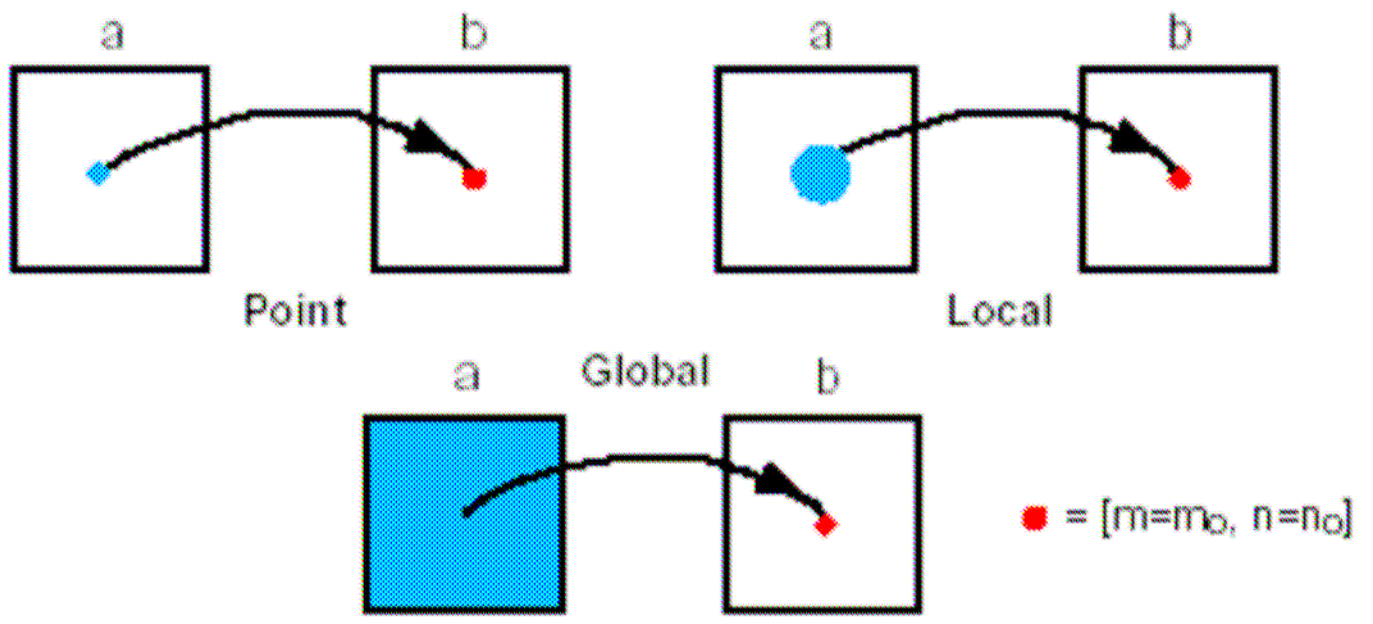

Figure 6 :Illustration of various types of image operation

#### 2.2.2.2 Types of neighborhoods:

Neighborhood operations play a key role in modern digital image processing. It is therefore important to understand how images can be sampled and how that relates to the various neighborhoods that can be used to process an image.

• Rectangular sampling – In most cases, images are sampled by laying a rectangular grid over an image as illustrated in Figure 5. This results in the type of sampling shown in Figure 7c.

• Hexagonal sampling – An alternative sampling scheme is shown in Figure 7b and is termed hexagonal sampling. Local operations produce an output pixel value $b[m=m_o,n=n_o]$ based upon the pixel values in the *neighborhood* of $a[m=m_o,n=n_o]$. Some of the most common neighborhoods are the 4-connected neighborhood and the 8-connected neighborhood in the case of rectangular sampling and the 6-connected neighborhood in the case of hexagonal sampling illustrated in Figure 7.

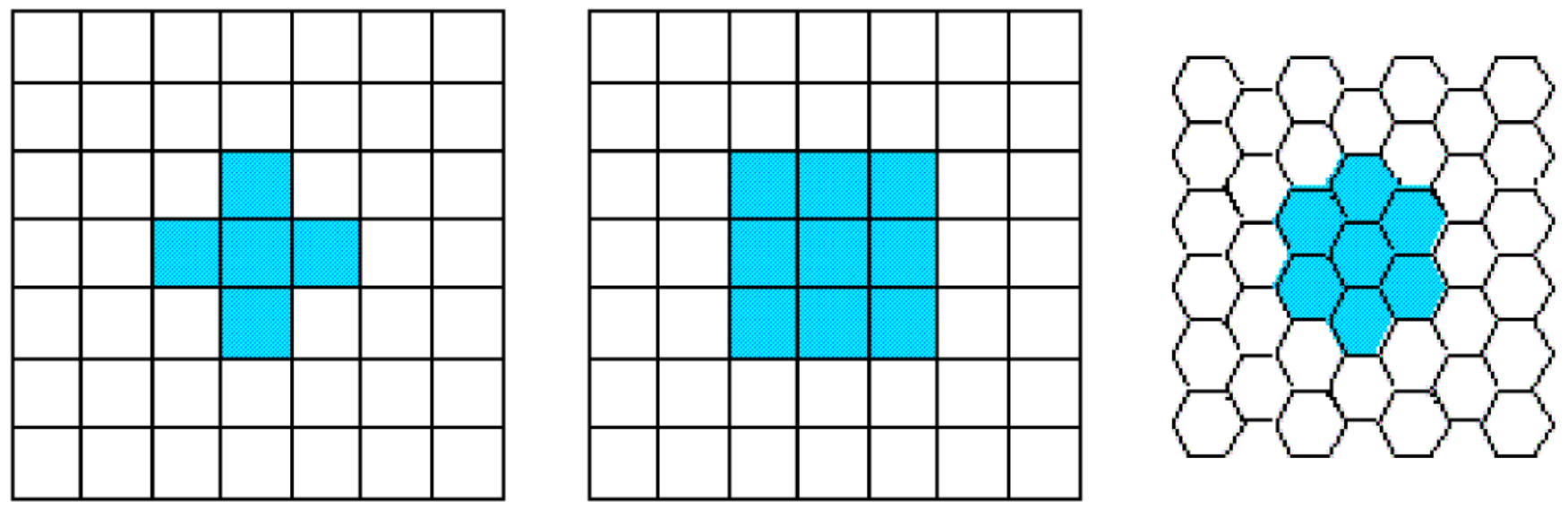

Figure 7a:Rectangular sampling 4 connected 7b:Rectangular Sampling 8-connected 7c:hexagonal sampling 6-connected

### 2.2.3 Video Parameters:

It is appropriate—given that many static images are derived from video cameras and frame grabbers— to mention the standards that are associated with the three standard video schemes that are currently in worldwide use – NTSC,PAL, and SECAM. This information is summarized in Table 2.3.

| *Standard* / *Property* | NTSC | PAL | SECAM |
|---|---|---|---|
| images / second | 29.97 | 25 | 25 |
| ms / image | 33.37 | 40.0 | 40.0 |
| lines / image | 525 | 625 | 625 |
| (horiz./vert.) = aspect ratio | 4:3 | 4:3 | 4:3 |
| interlace | 2:1 | 2:1 | 2:1 |
| µs / line | 63.56 | 64.00 | 64.00 |

Table 2.3:Standard video parameters

In an interlaced image the odd numbered lines (1,3,5,…) are scanned in half of the allotted time (e.g. 20 ms in PAL) and the even numbered lines (2,4,6,…) are scanned in the remaining half. The image display must be coordinated with this scanning format. The reason for interlacing the scan lines of a video image is to reduce the perception of flicker in a displayed image. If one is planning to use images that have been scanned from an interlaced video source, it is important to know if the two half-images have been appropriately "shuffled" by the digitization hardware or if that should be implemented in software. Further, the analysis of moving objects requires special care with interlaced video to avoid "zigzag" edges. The number of rows ($N$) from a video source generally corresponds one–to–one with lines in the video image. The number of columns, however, depends on the nature of the electronics that is used to digitize the image. Different frame grabbers for the same video camera might produce $M$ = 384, 512, or 768 columns (pixels) per line.

**2.3 Typical Problems**

- **Geometric transformations** such as enlargement, reduction, and rotation.
- **Color corrections** such as brightness and contrast adjustments, quantization, or conversion to a different color space
- **Registration** (or alignment) of two or more images
- **Combination** of two or more images, e.g. into an average, blend, difference, or image composite
- **Interpolation, demosaicing**, and **recovery** of a full image from a RAW image format like a BAYER filter pattern.
- **Segmentation** of the image into regions

- **Image editing** and digital retouching.
- **Extending dynamic range** by combining differently exposed images (generalized signal averaging of Wyckoff sets)

Besides static two-dimensional images, the field also covers the processing of time-varying signals such as video and the output of tomographic equipment. Some techniques, such as morphological image processing, are specific to binary or grayscale images.

The procedure explained in our report is a monocular approach for estimation of depth of a scene. The idea involves mapping the Pixel Depth of the object photographed in the image with the Real Depth of the object from the camera lens with an interpolation function. In order to find the parameters of the interpolation function, a set of lines with predefined distance from camera is used, and then the distance of each line from the bottom edge of the picture (as the origin line) is calculated.

### 3.1 Pixel Depth

The Pixel Depth of an object in an image is defined as the distance of the foot of the object from the edge of the photograph of the same object. Since the distances are measured from the top of the screen in a computer, the pixel depth in accordance with the adjoining picture can be calculated as

$$\text{pixel depth} = R - Ri$$

where

R : the size of the image

Ri: the distance of the foot of the object from the periphery of the image

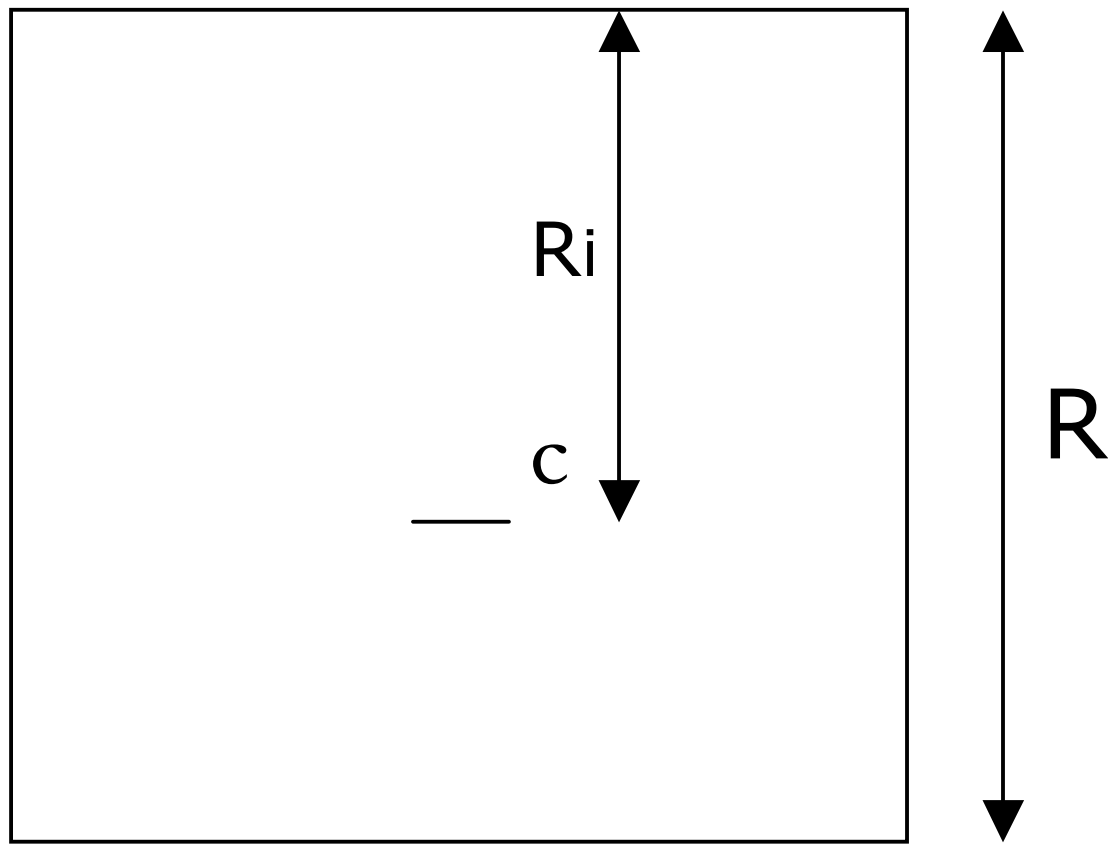

Figure 8: The Pixel Depth

### 3.2 Real Depth

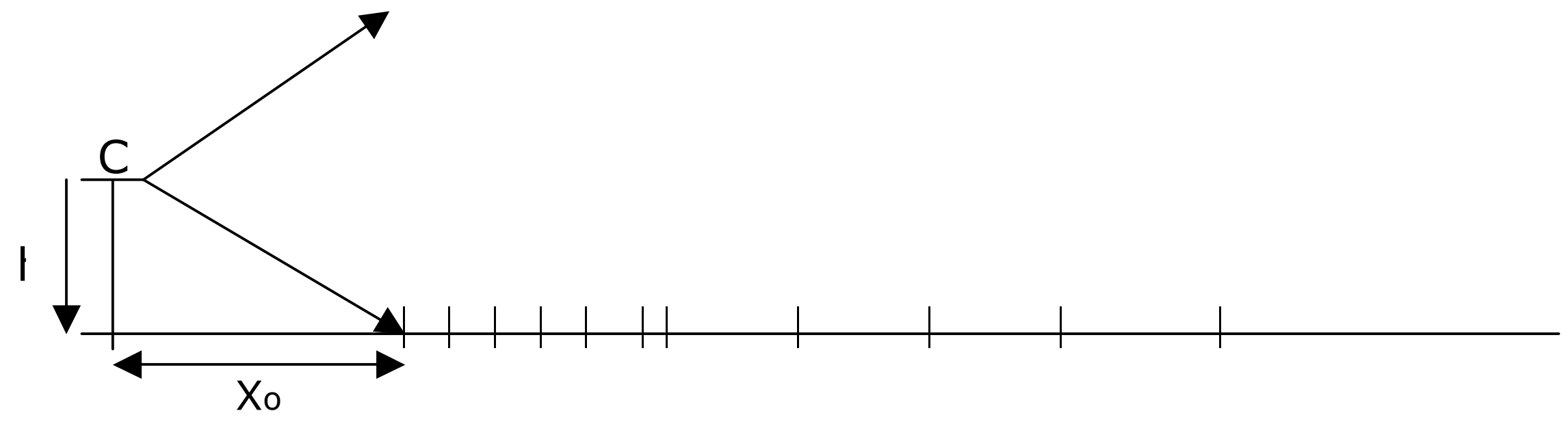

Figure 9 : The Real Depth

The Real Depth of an object is defined as the horizontal distance between the camera lens and the object being photographed. The camera (C) is maintained in a horizontal orientation at a height 'h' above the ground. The rays emanating out of the camera represent the region of vision of the camera. The point Xo is the distance from the foot of the camera beyond which the image is visible.

The distance of first sight on the ground, defined by Xo, is related to the height 'h' of the camera lens above the ground. If the height of the camera above the ground is increased, Xo also increases, as is shown in the picture below

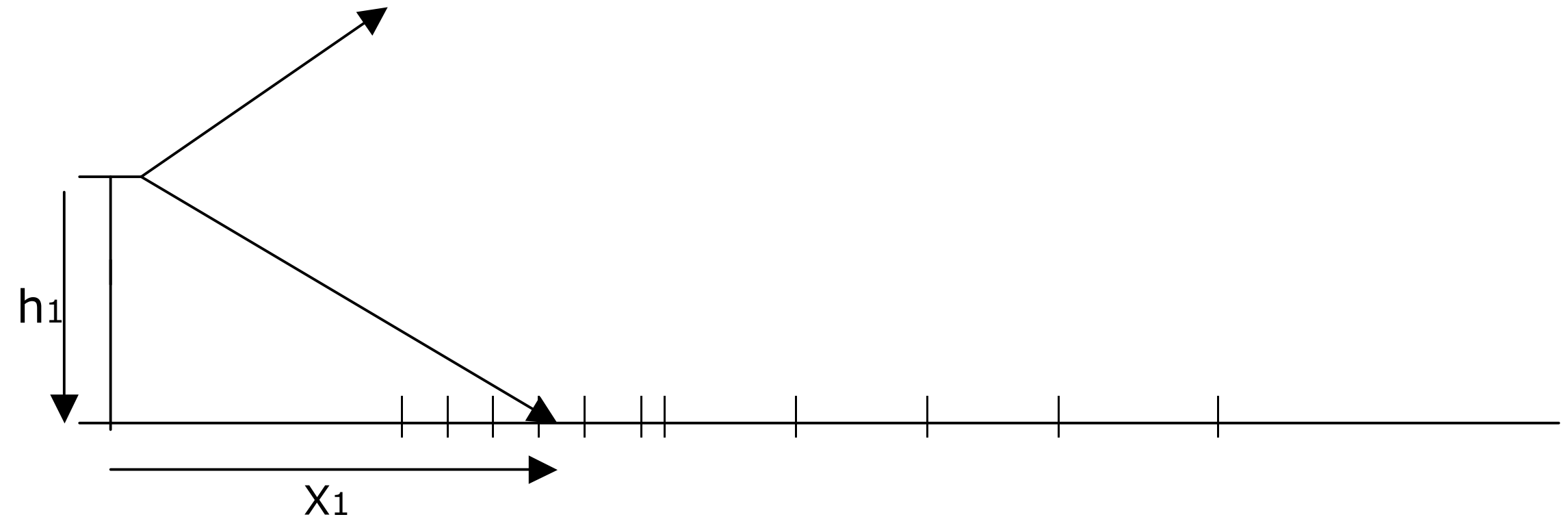

Figure 9 : Variation of 'Xo' with 'h'

### 3.3 Interpolation

This method includes two steps :

First, calculating an interpolation function based on the height and the horizontal angle of the camera. Second, using this function to calculate the distance of the object from the camera.

In the first step, named the **primitive evaluation phase**, the camera is located in a position with a specified height and absolutely horizontal with respect to the ground. Then from this position, we take a picture from some lines with equal distance from each other. Then, we provide a table in which the first column is the number of pixels counted from each line to the bottom edge of the captured picture (as the origin line), and the second column is the actual distance of that line from the camera position. Now, by assigning an interpolation method to this table, the related interpolation polynomial is calculated

$$y = \sum_{n=0}^{M} a_n x^n$$

In this formula, **x** is the distance of the object from the camera, and **M** is the order of the polynomial in the evaluation environment in the first step. In the second step of this method - with the same height and horizontal angle of the camera - the number of the pixels between the bottom edge of the target in the image (the nearest edge of an object in the image to the base of the camera) and the bottom edge of the captured image is counted and considered as **x** values in the interpolation function.

The output of this function will be the real distance between the target image and the camera.

The output of this function will be the real distance between the target in the image and the camera.

**This method has some advantages in comparison to the previous methods:**

a) Using only a single camera for the depth finding.

b) Having no direct dependency on the camera parameters like focal length and etc.

c) Having uncomplicated calculations.

d) Requiring no auxiliary devices.

e) Having a constant response time, because of having a fixed amount of calculations; so it will be reliable for applications in which the response time is important.

f) This method can be used for both stationary and moving targets.

**However, This method has some limitations such as:**

a) The dependency on the camera height and horizontal angle, so that if both or one of them is changed, there will be a need to repeat the first step again.

### 3.4 The N – Degree Polynomial

As has been understood from the concept of interpolation, it is desired to devise a polynomial which can takes in as input the pixel depth of an object in an image and gives as output, the actual distance between the object and the camera lens, the real depth. The derivation of the polynomial simply implies the calculation of the coefficients of the various terms of $x^n$ in the polynomial. Once the polynomial has been completely derived, that is, all the coefficients are known, the polynomial in itself becomes an intelligent system to calculate the real depth of a particular image if the pixel depth is provided.

This polynomial is unique for a particular camera and a particular height. Altering the focus of the camera or the height requires that the polynomial be derived again for the new orientation.

The core of our project lies in deriving the polynomial which can automatically calculate the Real Depth of the image. In our initial approach towards the derivation of the interpolation polynomial, we assumed that the height of the camera above the ground to be constant and the camera be placed horizontally with respect to the ground. That is, there be no angle of inclination between the camera plane and the ground taken as reference.

In the initial set up for recording the pictures, the camera was positioned on an elevated platform above the ground and this height was measured. This height is called 'h'.

For calculation of the polynomial coefficients, it is required to have a sample set constituting of sample images taken in such a manner that the distance of object ( human in our case ) from the camera is constantly varying in each image. This sample set of images is then analyzed and the Pixel Depth of the image is tabulated against the Real Depth that was measured while taking the pictures with the help of a measuring tape.

This sample set of about fifteen clicked images is then processed with the Image Processing Tool in MATLAB. The recorded observations are given as input into MATLAB and with the help of a curve fitting tool, a curve is obtained for the particular set of observations. This curve obtained is unique to the particular camera from which the pictures had been clicked and for the constant height. Any further pictures taken from that camera can now be directly analyzed from this curve.

Further, we applied polynomial approximation to the curve and obtained an N-

degree polynomial for the set of values. The curve fitting MATLAB tool automatically calculates the coefficients and hence, the polynomial has been derived.

Refer to **APPENDIX A** for the set of observation values and MATLAB images pertaining to the initial recordings taken.

Having analyzed the interpolation method for a particular constant height ‘h’ of the camera above the ground, we further analyzed the interpolation method for varying values of the height. Two more sets of observations were taken, for constantly increasing values of height.

As has been studied initially, increasing the values of height also has an effect on the closest point of sight, which is denoted by Xo. For each of the successive values of height ‘h’, a set of fifteen pictures were taken and each set was analysed as done in the initial set up. With the MATLAB image processing tool, each set of the sample values were analyzed and with the curve fitting tool, a polynomial was derived for each set of images obtained.

Refer to **APPENDIX B** for the set of observation values and MATLAB images pertaining to the initial recordings taken.

## RESULT

A set of fifteen images each was taken for three successively increasing values of height 'h'. These have been shown in **APPENDIX A , B and C.** Corresponding to each set of pictures, the closest observable point on the ground, which is denoted by Xo was found to be different, as was depicted in the previous section.

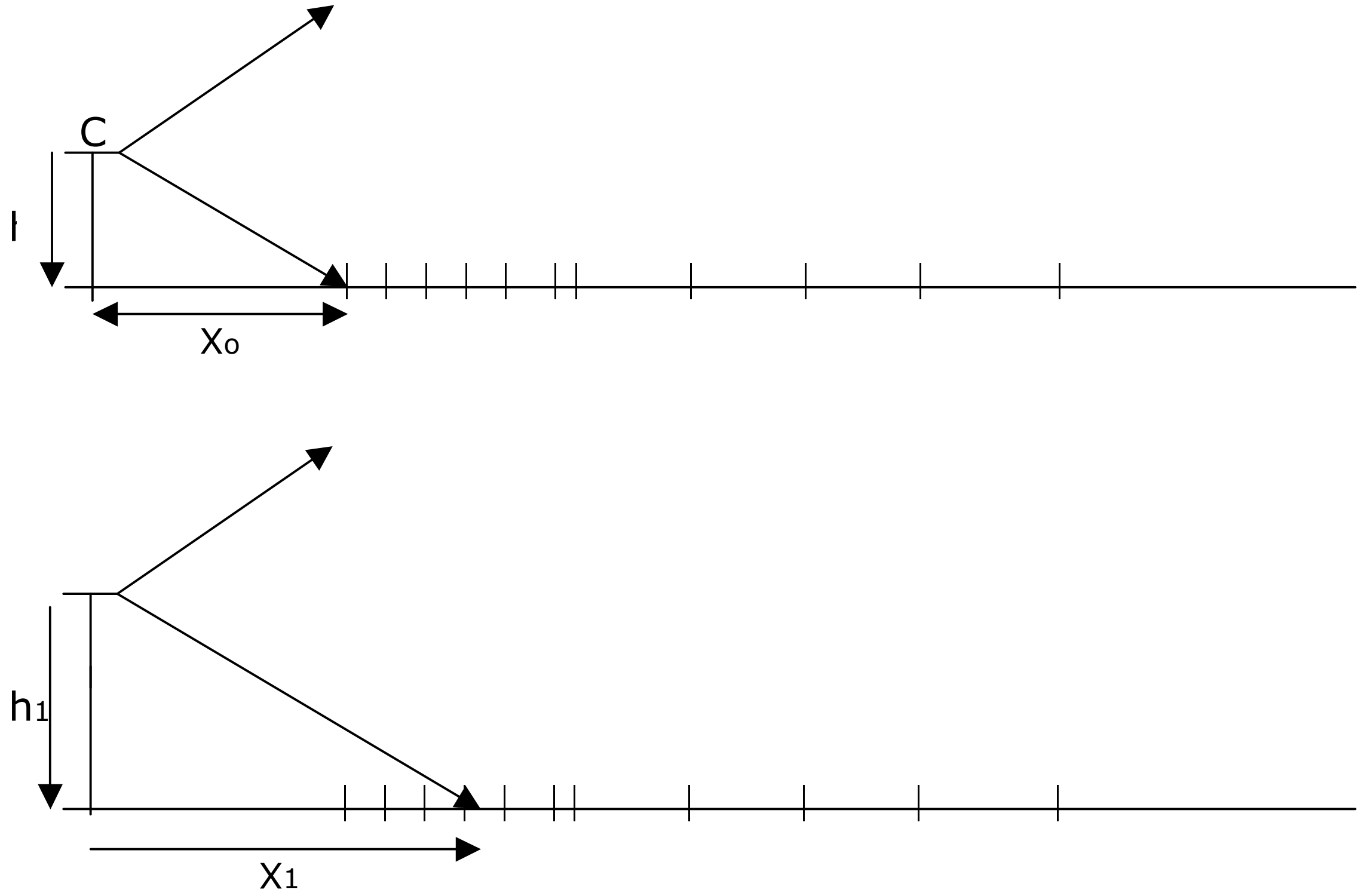

Variation of the Xo with the successive increase in the height 'h'

For each of the set of values, the sample set was analyzed and with the help of curve fitting tool, a particular polynomial approximation was achieved. This polynomial was decided on the basis of a comprehensive study of the various possible options available within the polynomials on the basis of the least error calculated in each of them.

## APPLICATIONS

Depth Estimation is a highly useful tool in the 3D structuring of a scene from a two dimensional image. Thus, it can be said that it provides a 2D to 3D mapping of images. This is highly instrumental since it can be applied to intelligent machines such as robots. Robots usually require the three dimensional understanding of the environment they are in to take decisions. This method can therefore be used in robots to provide the robot with a 3D sense of the scene. With the images perceived by the robot with its cameras, depth can be calculated with the help of interpolation polynomial and hence, the robot can decide the action that it has to take in case there is an obstacle in its path of movement.

This method can also be used to calculate the velocity of a moving object. With a system that can continuously monitor the moving object and can take pictures of the object constantly, this interpolation method can be applied to the set of images constantly being viewed and the velocity can be calculated. The accuracy of the speed estimation will depend on the rate at which the photos are clicked by the system.

The method can also be applied in cars to measure the distance between any two cars and hence can be incorporated as a safety algorithm to apply brakes automatically if the need be.

## FUTURE SCOPE

So far, while taking pictures, we assumed that the various parameters like focus of the camera, zoom, angle of the camera with respect to the ground, object movement, lens translation etc. remain constant throughout. But this doesn't always happen. Sometimes, errors may be induced in the approximation if any of these parameters is not constant or is present while it is assumed to be insignificant.

Taking the approach further would require the analysis of the effect of these parameters on the interpolation function. Here, in our process, we took matters a step further by increasing the value of 'h' for each sample space and then noting the variation in the interpolation function derived. Thus, we can say that a detailed and comprehensive study of this method would involve taking into account the other parameters that have been mentioned. Hence, we wish to develop this system further by incorporating these features into the algorithm design.

**APPENDIX A**

---

**Height of the camera above the ground, 'h' : 96.5 cms**

**Camera : SONY Digicam**

**Set at 5 Megapixel**

**Calculated Xo for this camera at the given h : 415 cms**

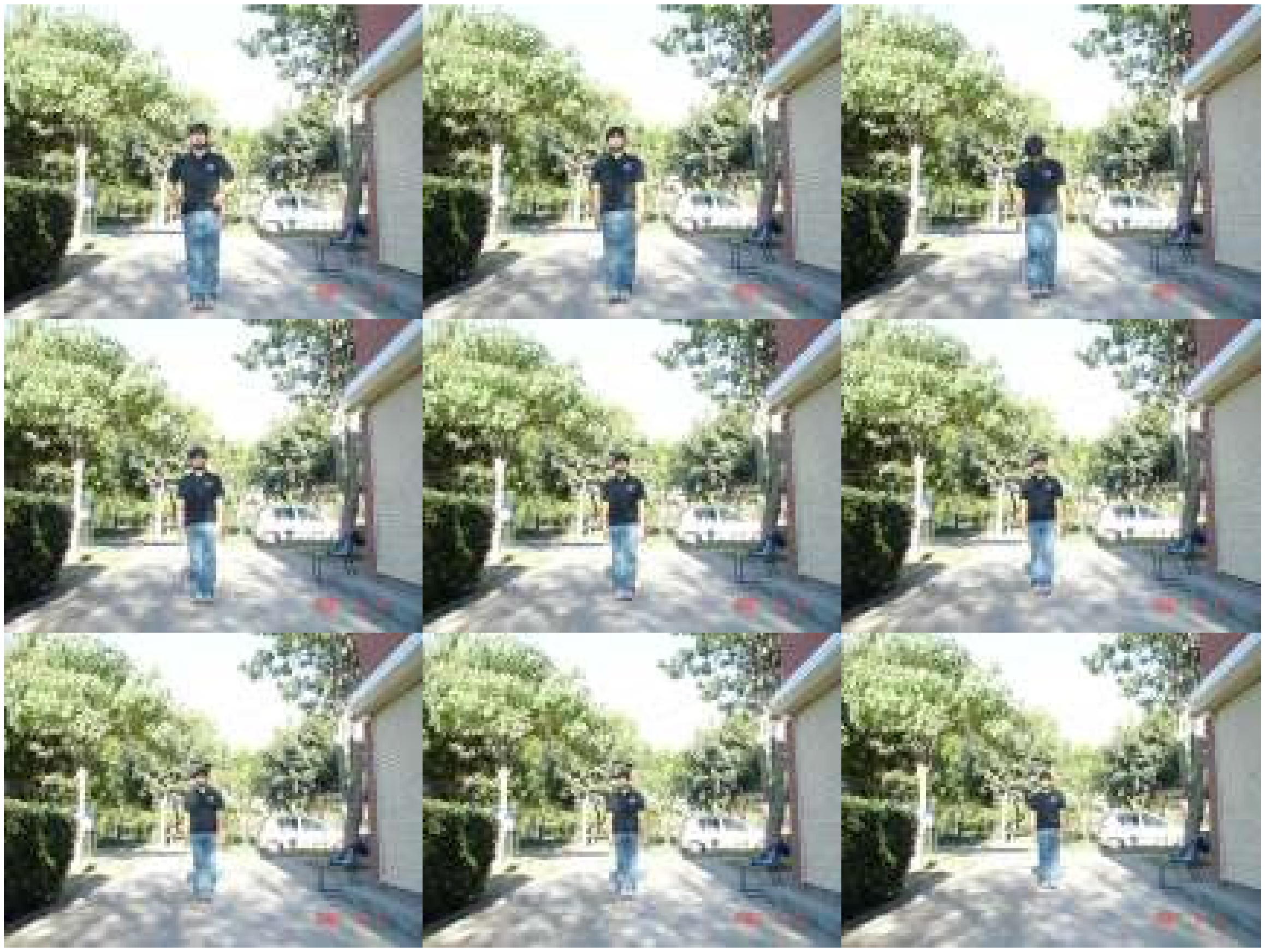

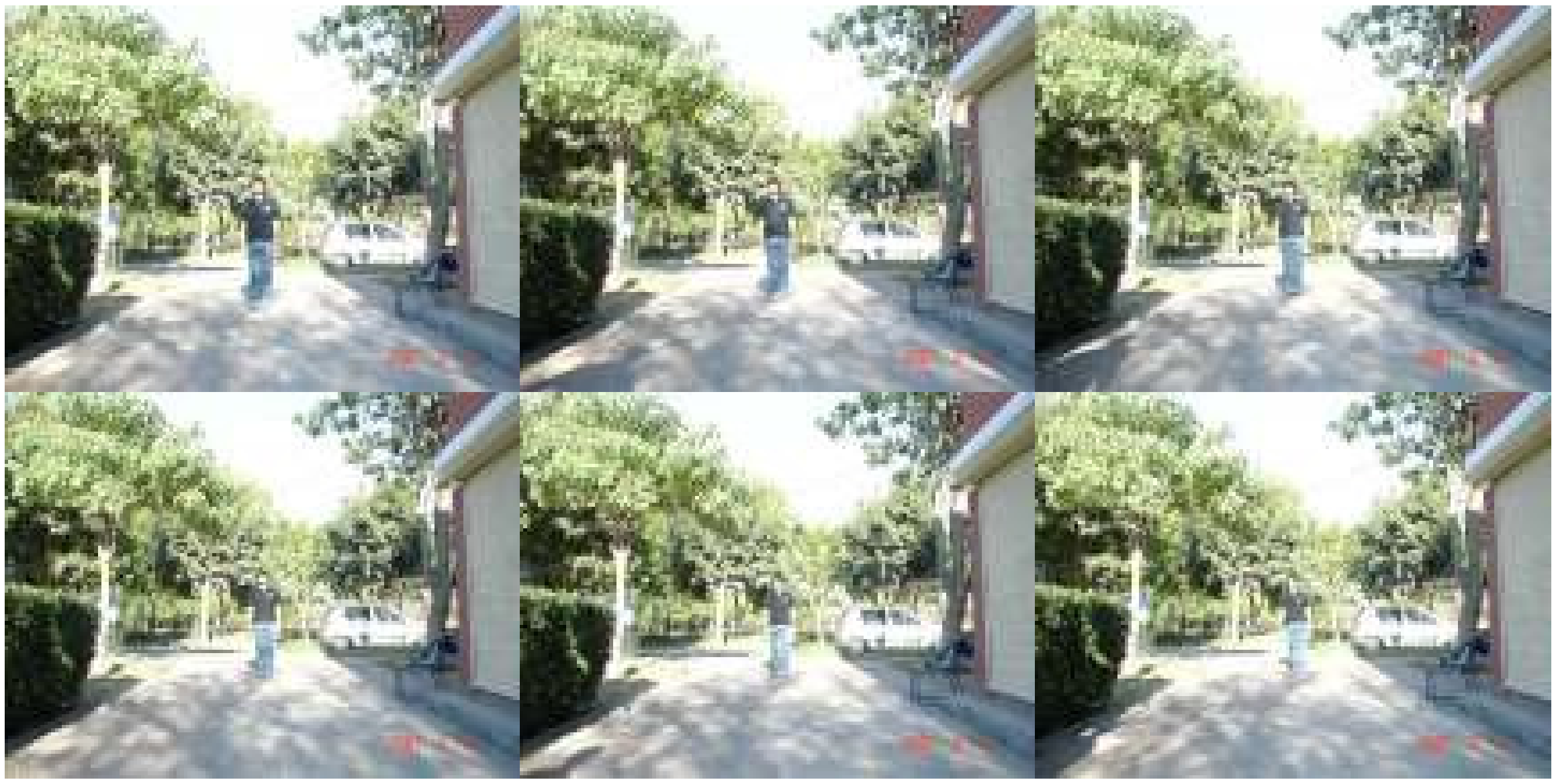

**Observations :**

| Photograph No. | Actual Depth (cm) | Pixel Depth (R-r) | R | r |
|---|---|---|---|---|
| 5711 | 450 | 55 | 1944 | 1889 |
| 5712 | 480 | 108 | 1944 | 1836 |
| 5713 | 510 | 136 | 1944 | 1808 |
| 5714 | 540 | 183 | 1944 | 1761 |
| 5715 | 570 | 202 | 1944 | 1742 |
| 5716 | 600 | 245 | 1944 | 1699 |
| 5717 | 630 | 287 | 1944 | 1657 |
| 5718 | 660 | 306 | 1944 | 1638 |
| 5719 | 720 | 365 | 1944 | 1579 |
| 5720 | 780 | 417 | 1944 | 1527 |
| 5721 | 840 | 458 | 1944 | 1486 |
| 5722 | 900 | 488 | 1944 | 1456 |
| 5723 | 960 | 507 | 1944 | 1437 |
| 5724 | 1020 | 513 | 1944 | 1431 |
| 5725 | 1080 | 534 | 1944 | 1410 |

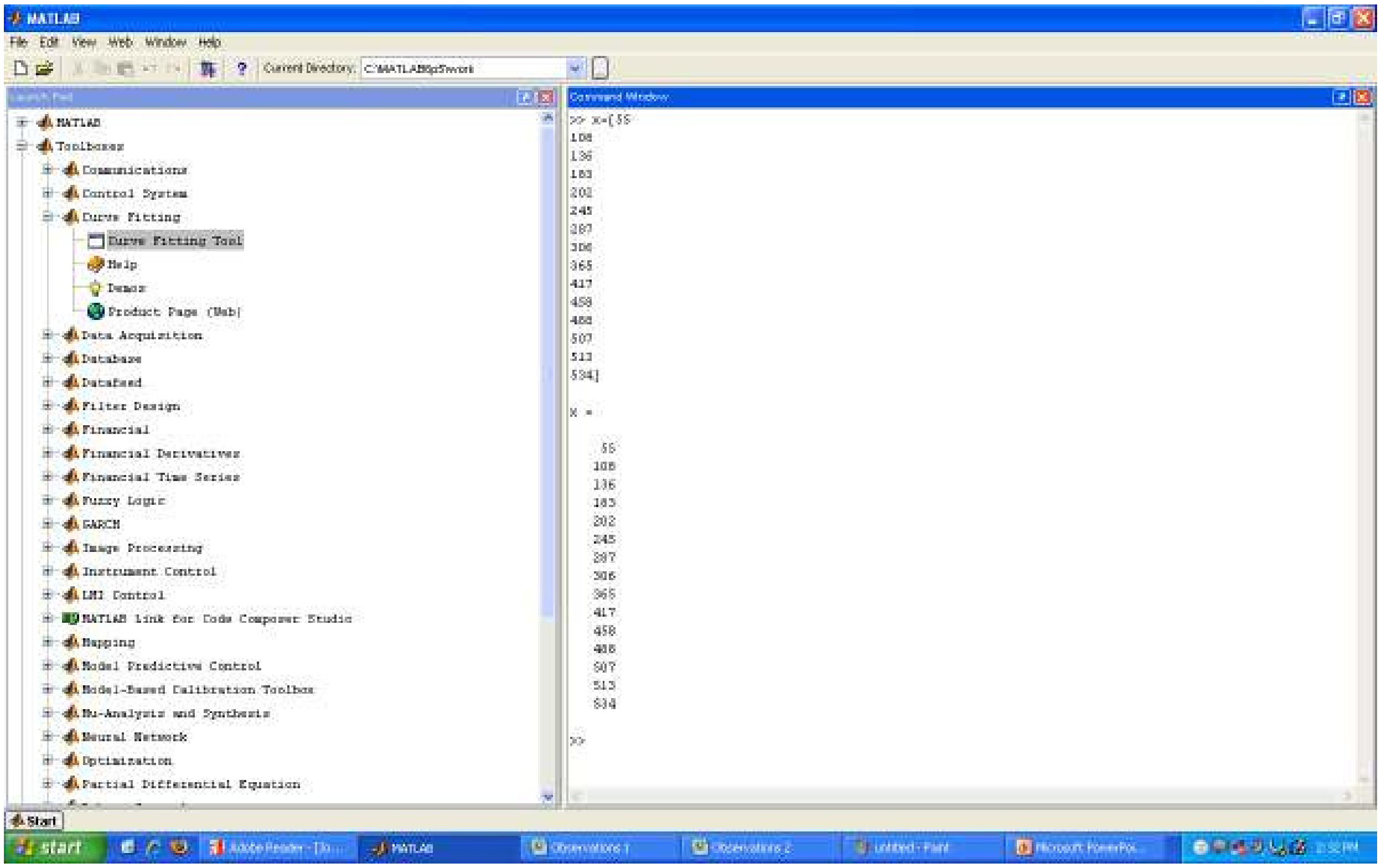

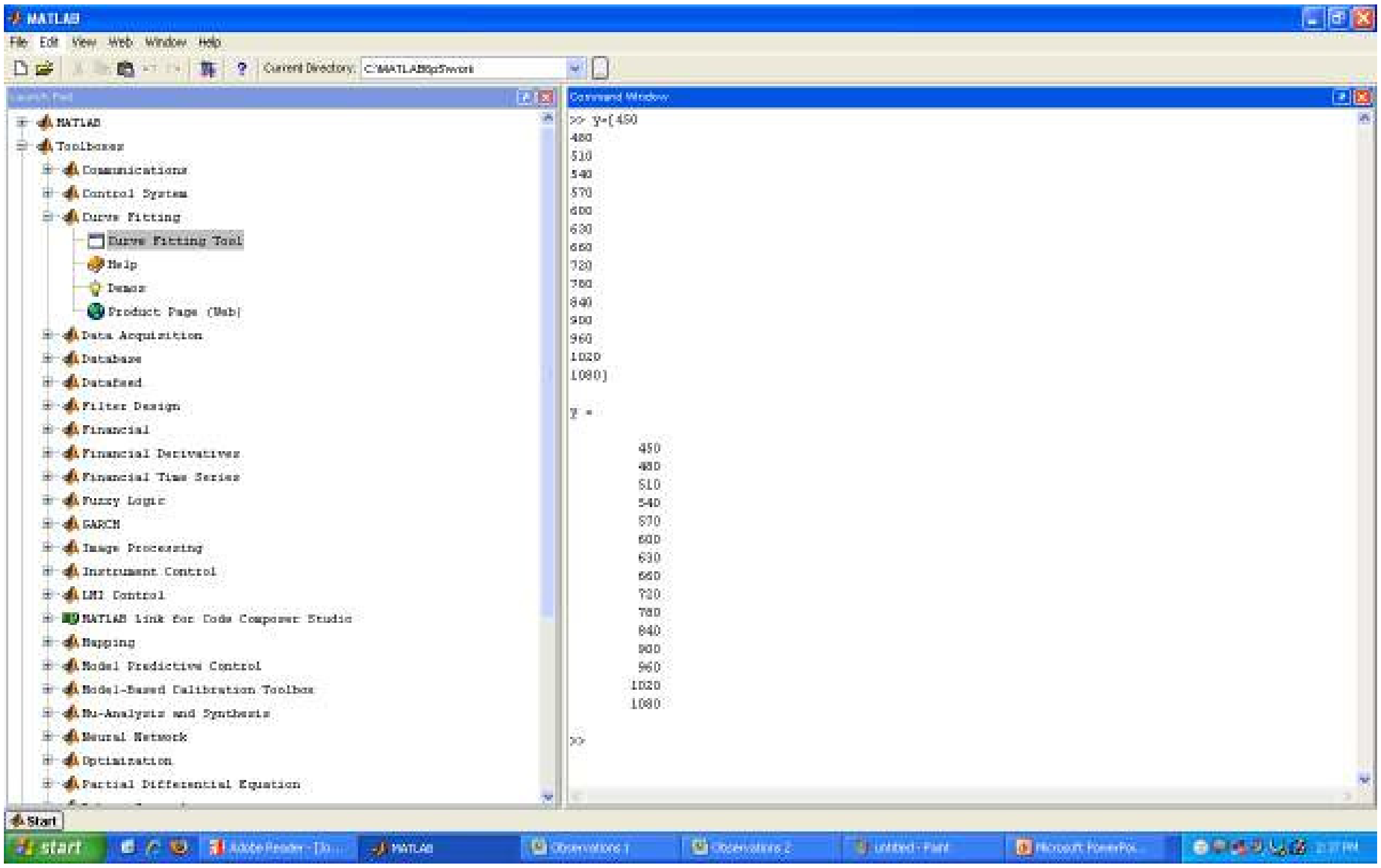

The following image shows the mapping of various values of x with the corresponding values of y. The mapping can take up any kind of a polynomial depending which suits it best which is decided by the error calculated in each approximation.

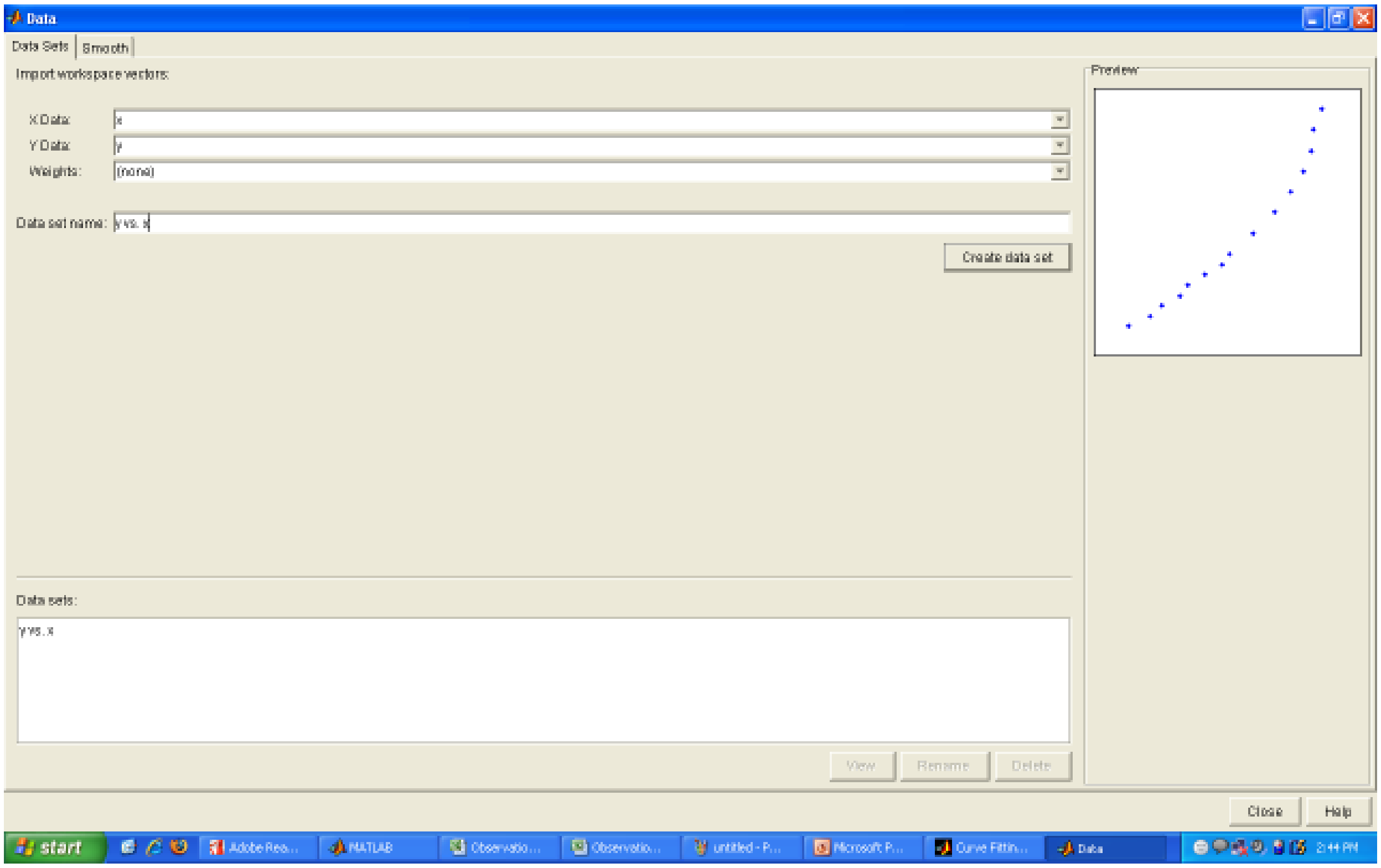

## 8th Degree Polynomial Approximation

After a series of successive examinations for the various options available in the Curve Fitting Tool, it was found that, for the mapping shown in the previous image, the 8th degree polynomial gives the best fit. That is, the root mean square error is minimum for this approximation. Hence, we get an 8th degree interpolation function as shown in the following image.

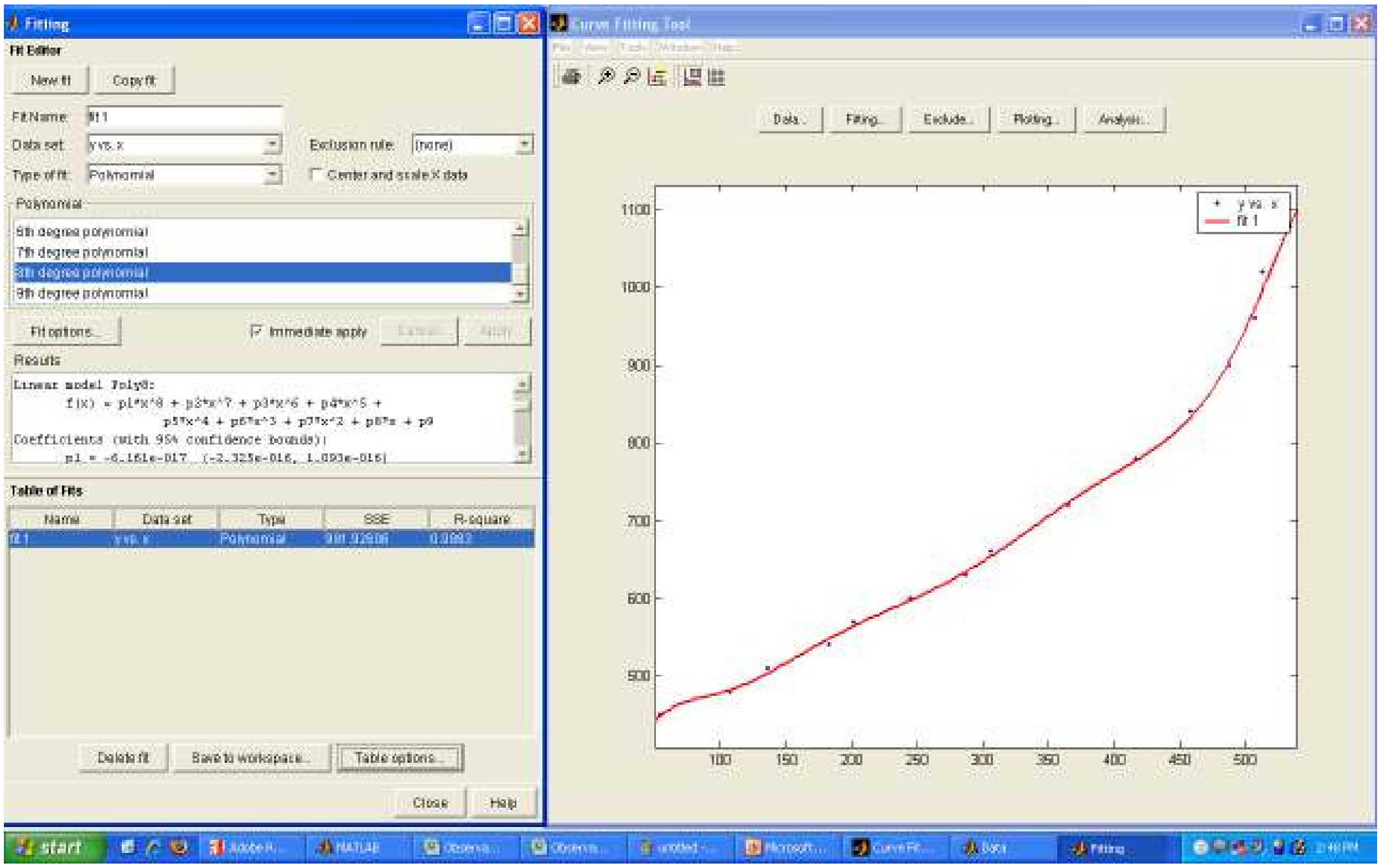

**Linear model Poly8:**

f (x) = p1*x^8 + p2*x^7 + p3*x^6 + p4*x^5 + p5*x^4 + p6*x^3 + p7*x^2 + p8*x + p9

**Coefficients:**

p1 = -6.161e-017 (-2.325e-016, 1.093e-016)
p2 = 1.427e-013 (-2.732e-013, 5.586e-013)
p3 = -1.372e-010 (-5.62e-010, 2.876e-010)
p4 = 7.095e-008 (-1.654e-007, 3.073e-007)
p5 = -2.142e-005 (-9.916e-005, 5.632e-005)
p6 = 0.003822 (-0.01149, 0.01913)
p7 = -0.3881 (-2.127, 1.351)
p8 = 20.93 (-81.14, 123)
p9 = 0.5922 (-2303, 2304)

**Goodness of fit:**

RMSE: 12.86

**APPENDIX B**

**Height of the camera above the ground, ‘h’ :  118 cms**

**Camera : SONY Digicam**

**Set at 5 Megapixel**

**Calculated Xo for this camera at the given h : 370 cms**

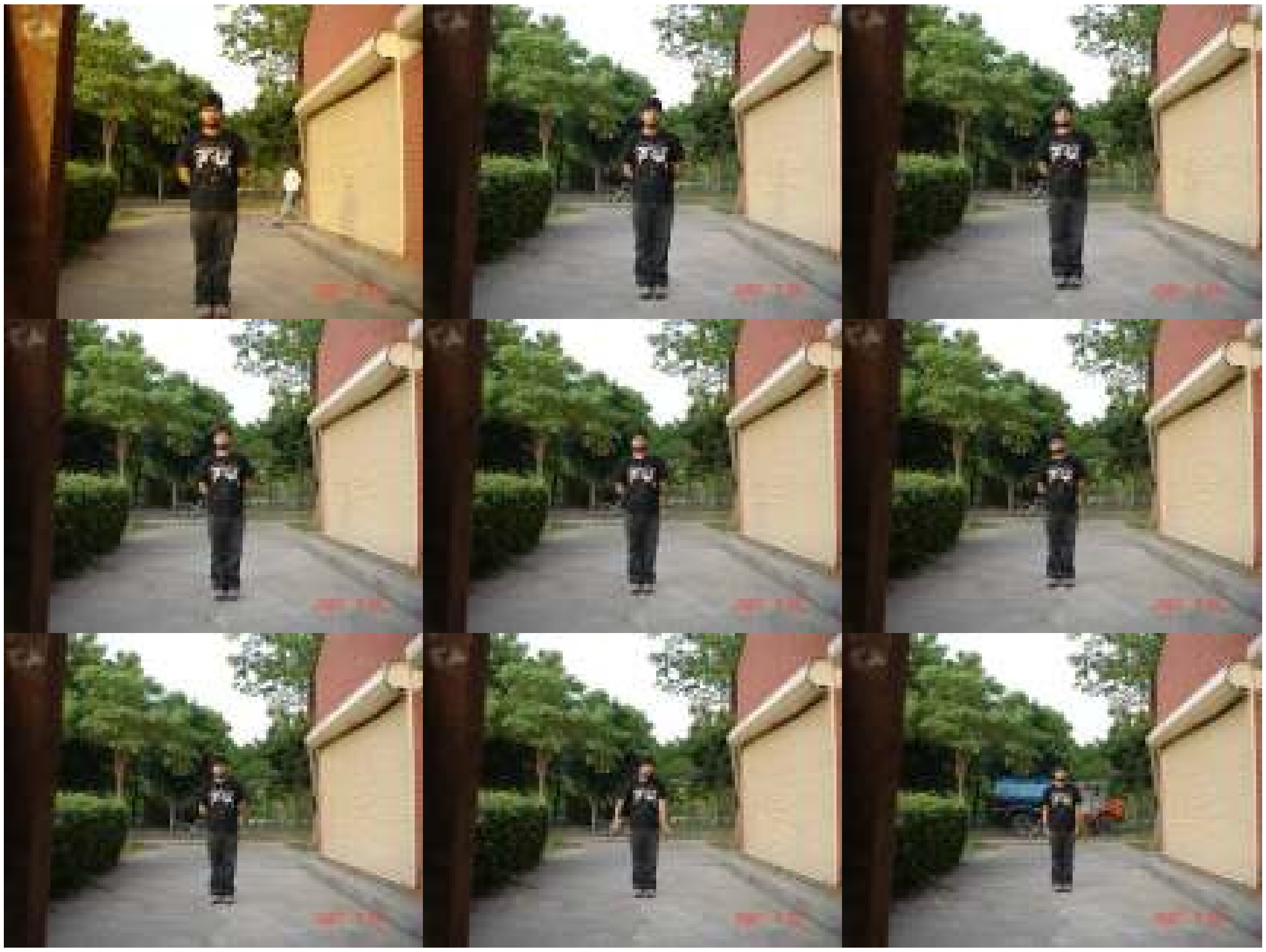

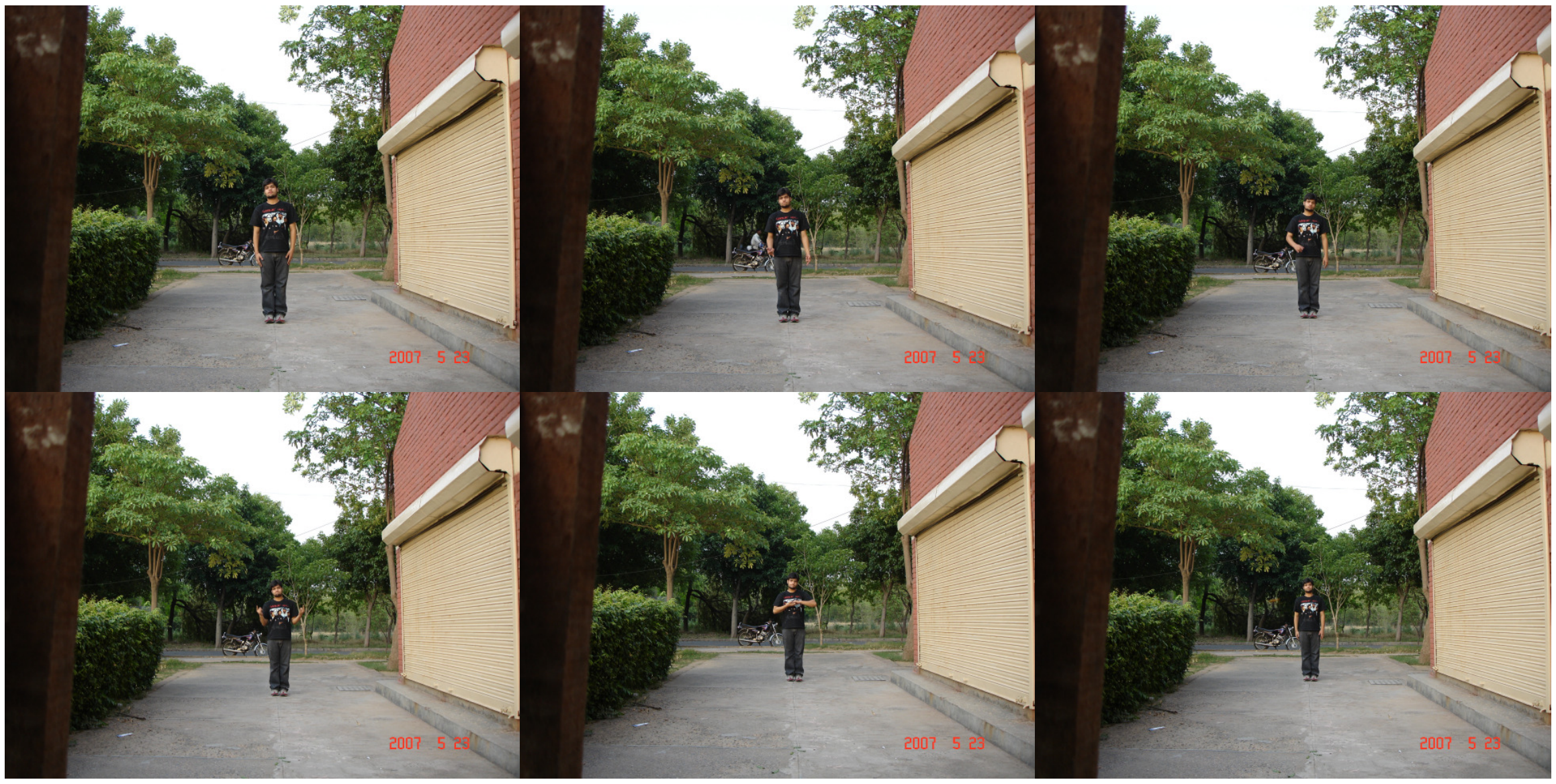

## Observations :

| Photograph No. | Actual Depth (cm) | Pixel Depth (R-r) | R | r |
|---|---|---|---|---|
| 6270 | 420 | 132 | 1944 | 1812 |
| 6271 | 450 | 202 | 1944 | 1742 |
| 6272 | 480 | 216 | 1944 | 1728 |
| 6273 | 510 | 258 | 1944 | 1686 |
| 6274 | 540 | 290 | 1944 | 1654 |
| 6275 | 570 | 303 | 1944 | 1641 |
| 6276 | 600 | 315 | 1944 | 1629 |
| 6277 | 660 | 353 | 1944 | 1591 |
| 6278 | 720 | 363 | 1944 | 1581 |
| 6279 | 780 | 376 | 1944 | 1568 |
| 6280 | 840 | 403 | 1944 | 1541 |
| 6281 | 900 | 419 | 1944 | 1525 |
| 6282 | 960 | 472 | 1944 | 1472 |
| 6283 | 1020 | 493 | 1944 | 1451 |
| 6284 | 1080 | 505 | 1944 | 1439 |

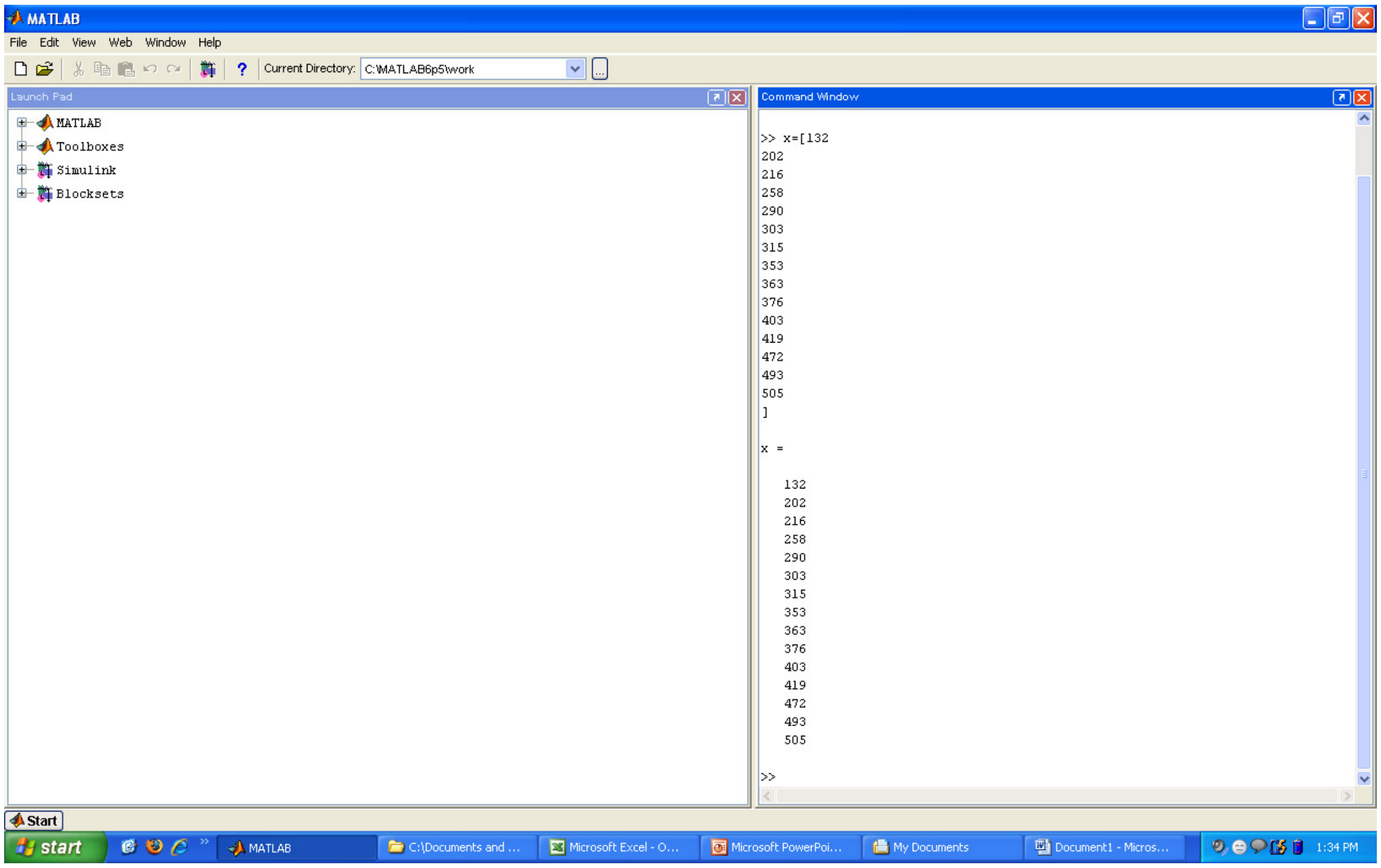

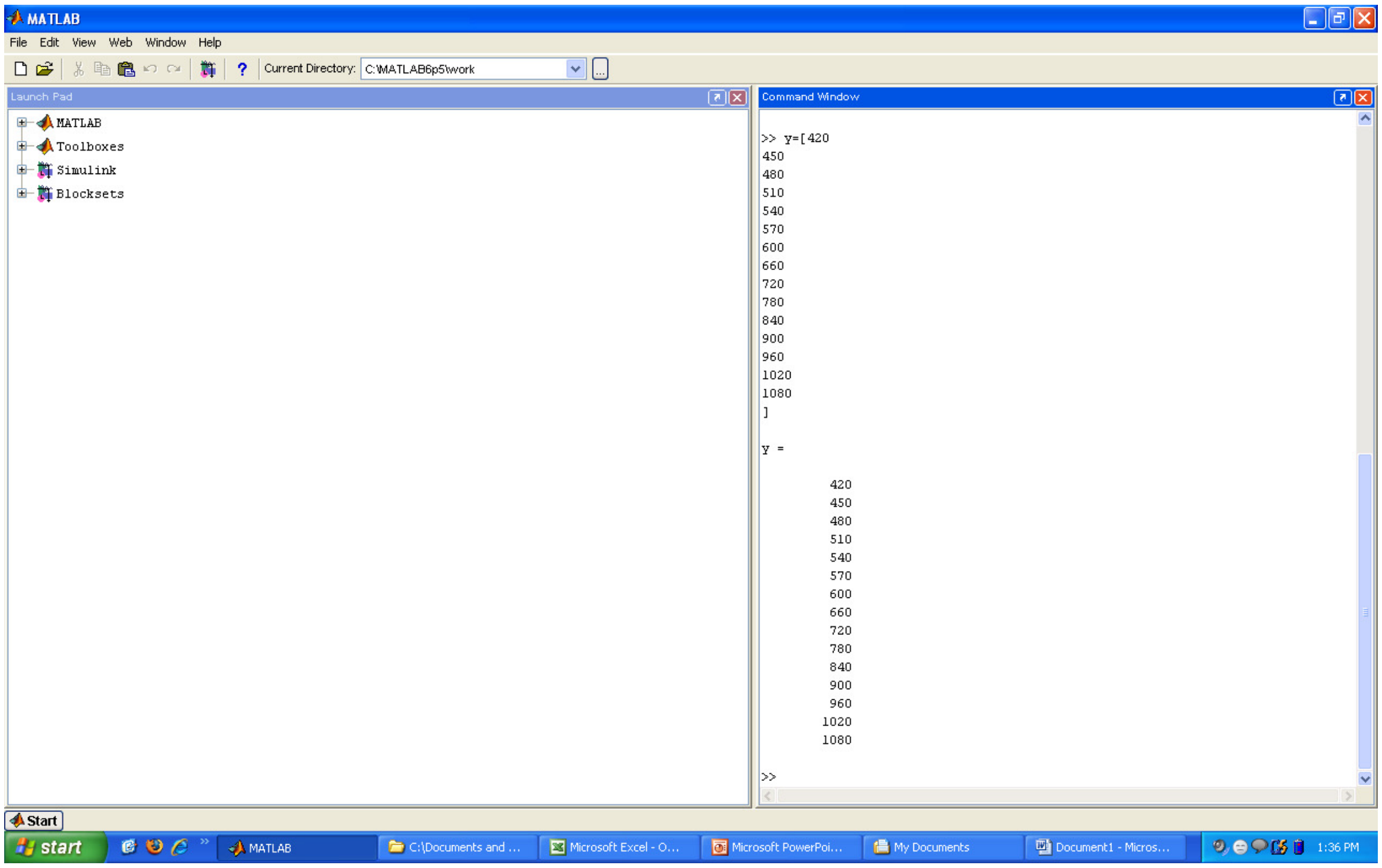

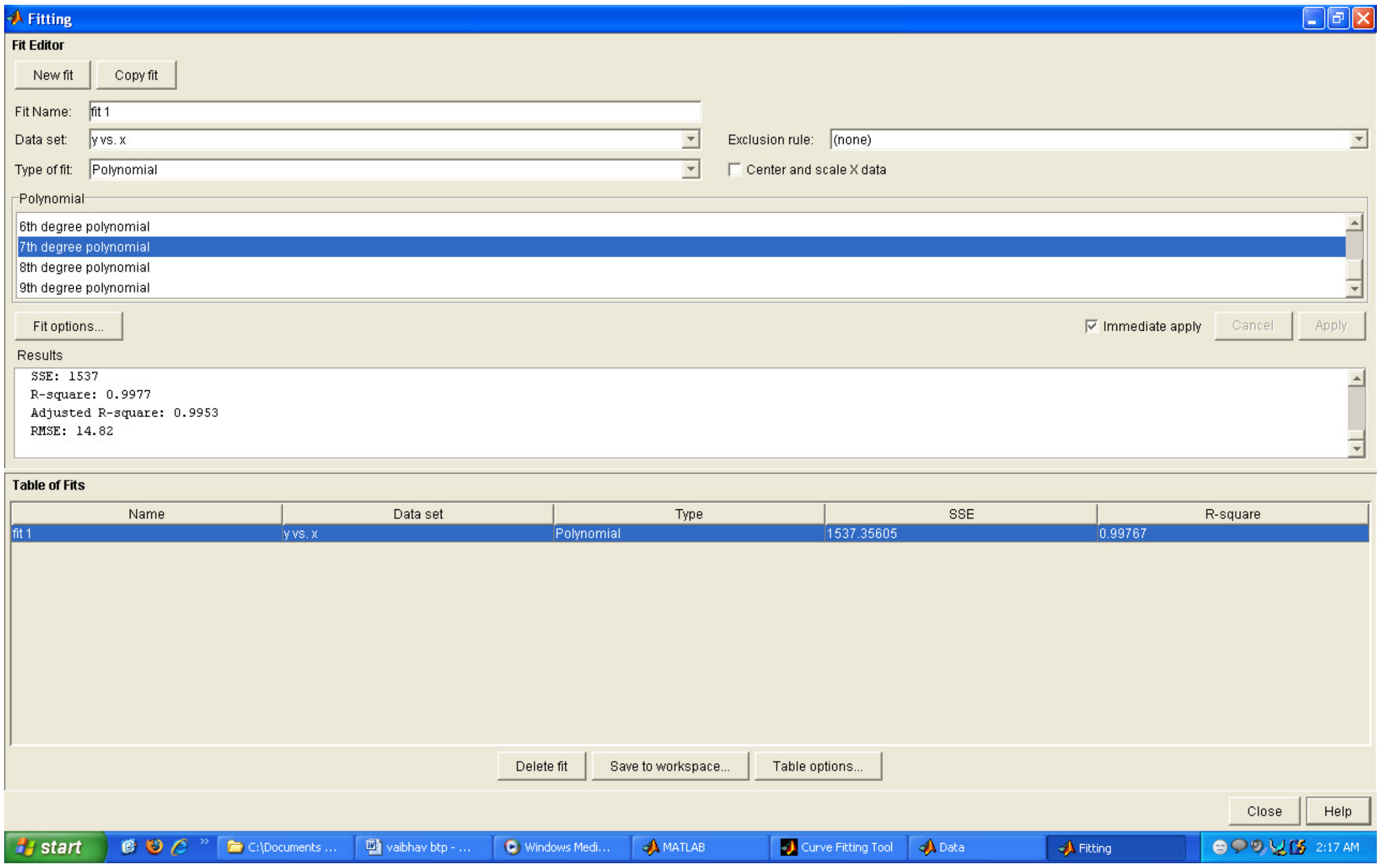

# 7th Degree Polynomial Approximation

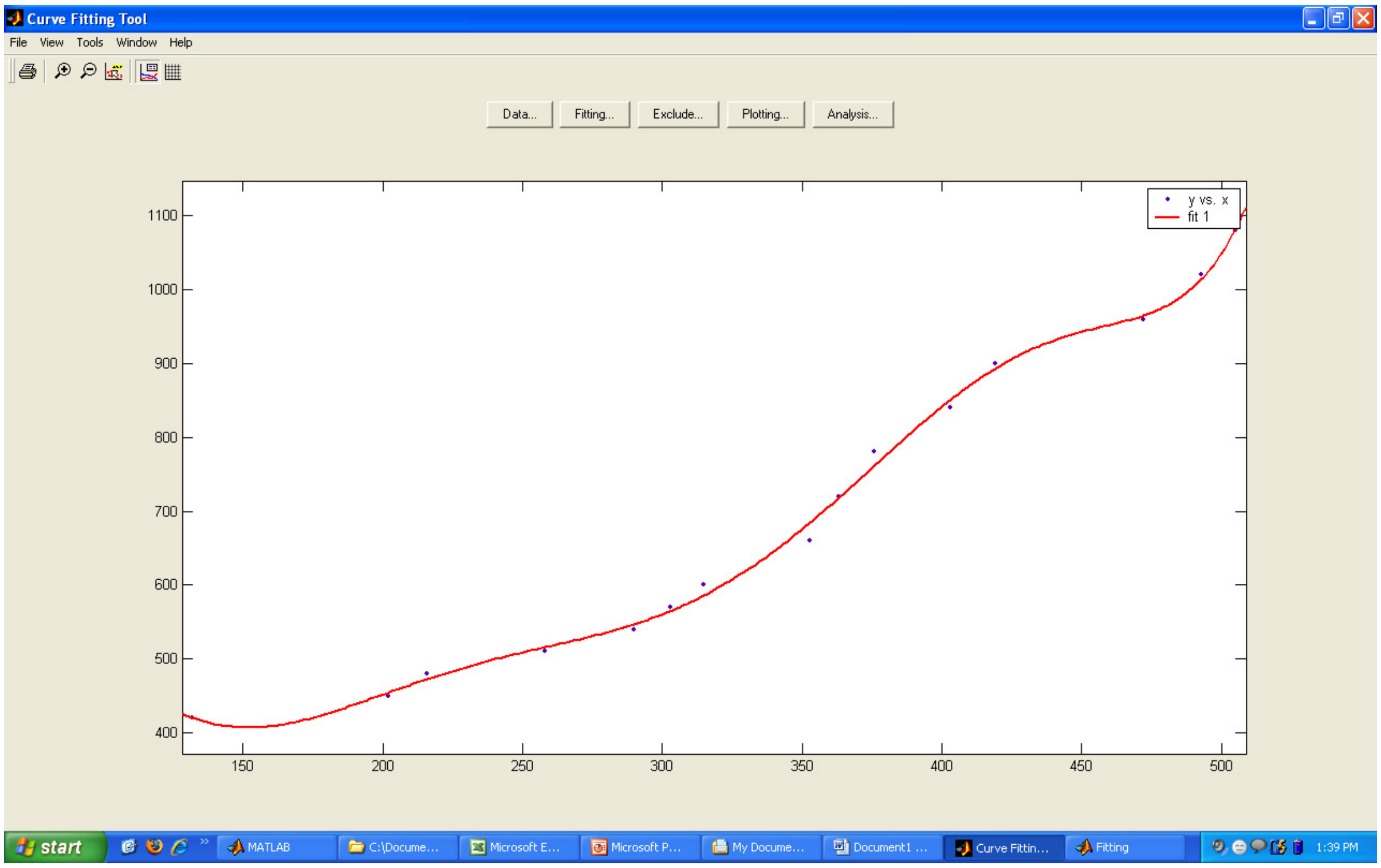

For these set of values, it was found that the 7th degree polynomial gives the least mean square error. Hence, interpolation was done using a 7th degree polynomial.

**Linear model Polynomial with degree 7:**

f(x) = p1*x^7 + p2*x^6 + p3*x^5 + p4*x^4 + p5*x^3 + p6*x^2 + p7*x + p8

**Coefficients:**

p1 = 3.499e-014 (-1.576e-013, 2.276e-013)
p2 = -6.005e-011 (-5.078e-010, 3.877e-010)
p3 = 3.837e-008 (-3.976e-007, 4.743e-007)
p4 = -1.028e-005 (-0.0002401, 0.0002195)
p5 = 0.0003892 (-0.07022, 0.07099)
p6 = 0.3663 (-12.24, 12.97)
p7 = -70.13 (-1275, 1135)
p8 = 4276 (-4.302e+004, 5.157e+004)

**Goodness of fit:**

RMSE: 14.82

## APPENDIX C

---

**Height of the camera above the ground, ‘h’ : 141.8 cms**

**Camera : SONY Digicam**

**Set at 5 Megapixel**

**Calculated Xo for this camera at the given h : 600 cms**

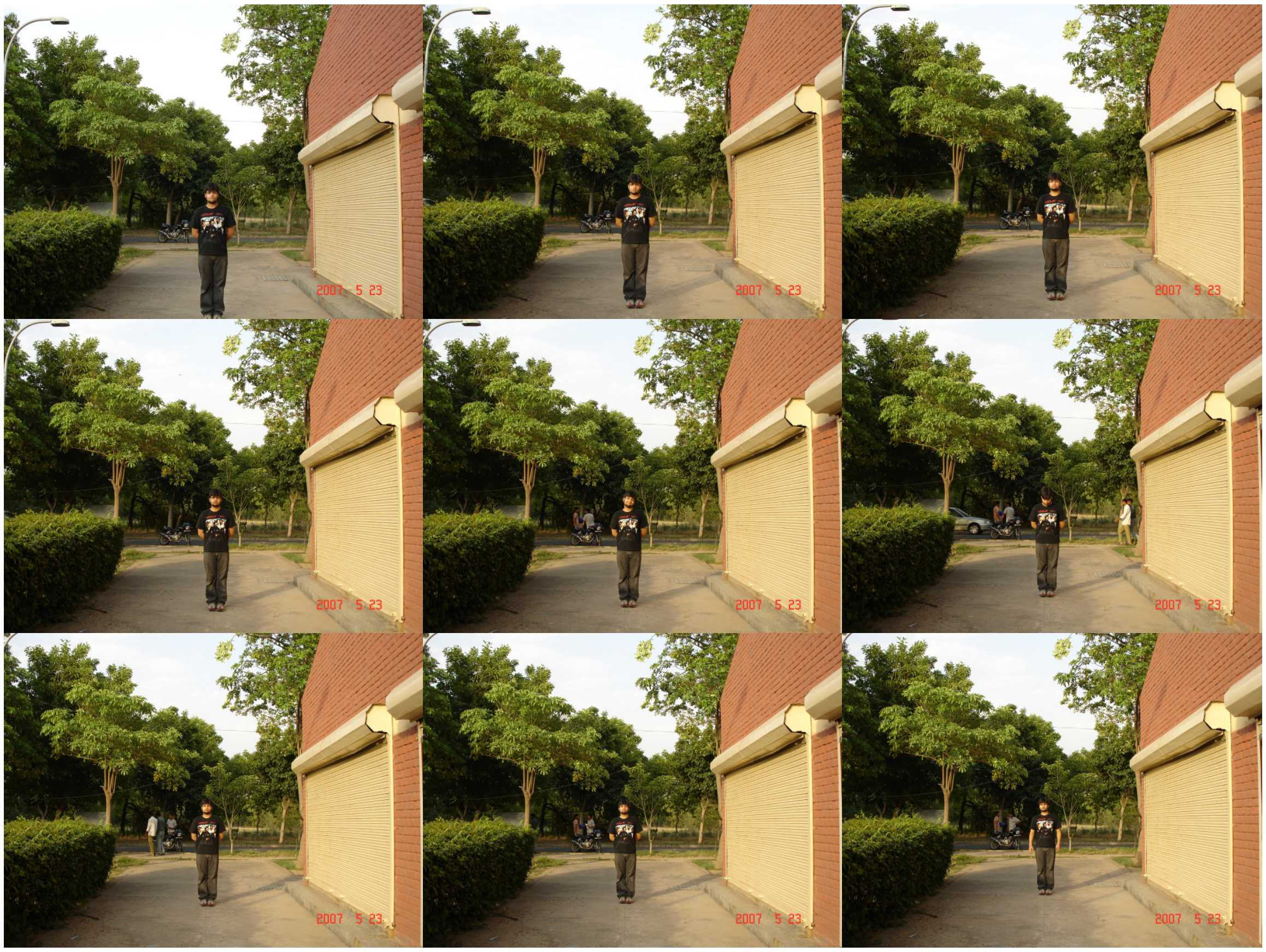

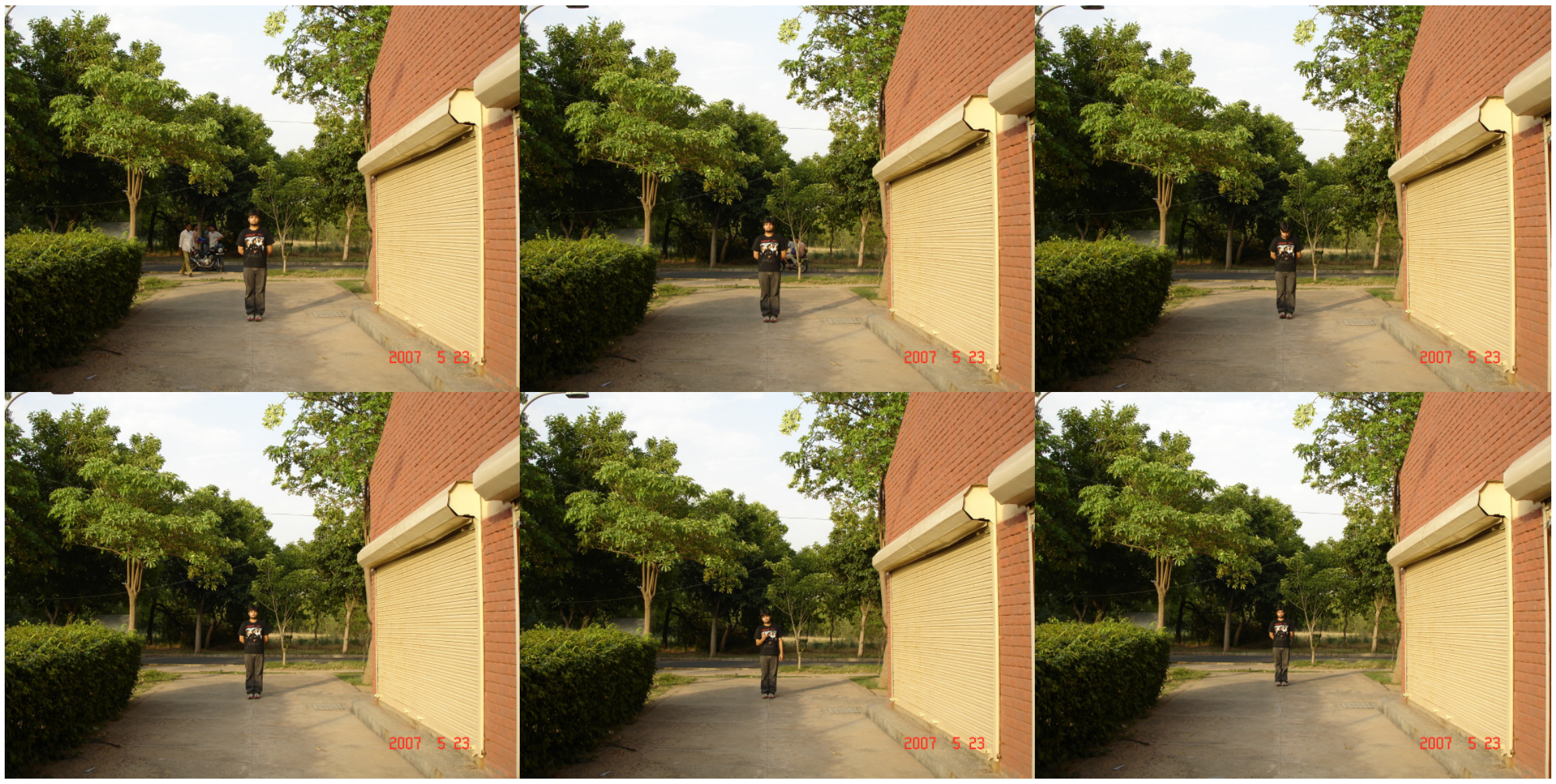

## Observations :

| Photograph No. | Actual Depth (cm) | Pixel Depth (R-r) | R | r |
|---|---|---|---|---|
| 6253 | 630 | 76 | 1944 | 1868 |
| 6254 | 660 | 125 | 1944 | 1819 |
| 6256 | 690 | 146 | 1944 | 1798 |
| 6257 | 720 | 171 | 1944 | 1773 |
| 6258 | 750 | 230 | 1944 | 1714 |
| 6259 | 780 | 262 | 1944 | 1682 |
| 6260 | 810 | 279 | 1944 | 1665 |
| 6261 | 870 | 328 | 1944 | 1616 |
| 6262 | 930 | 356 | 1944 | 1588 |
| 6263 | 990 | 363 | 1944 | 1581 |
| 6264 | 1050 | 374 | 1944 | 1570 |
| 6265 | 1110 | 393 | 1944 | 1551 |
| 6266 | 1170 | 406 | 1944 | 1538 |
| 6267 | 1230 | 437 | 1944 | 1507 |
| 6268 | 1290 | 469 | 1944 | 1475 |

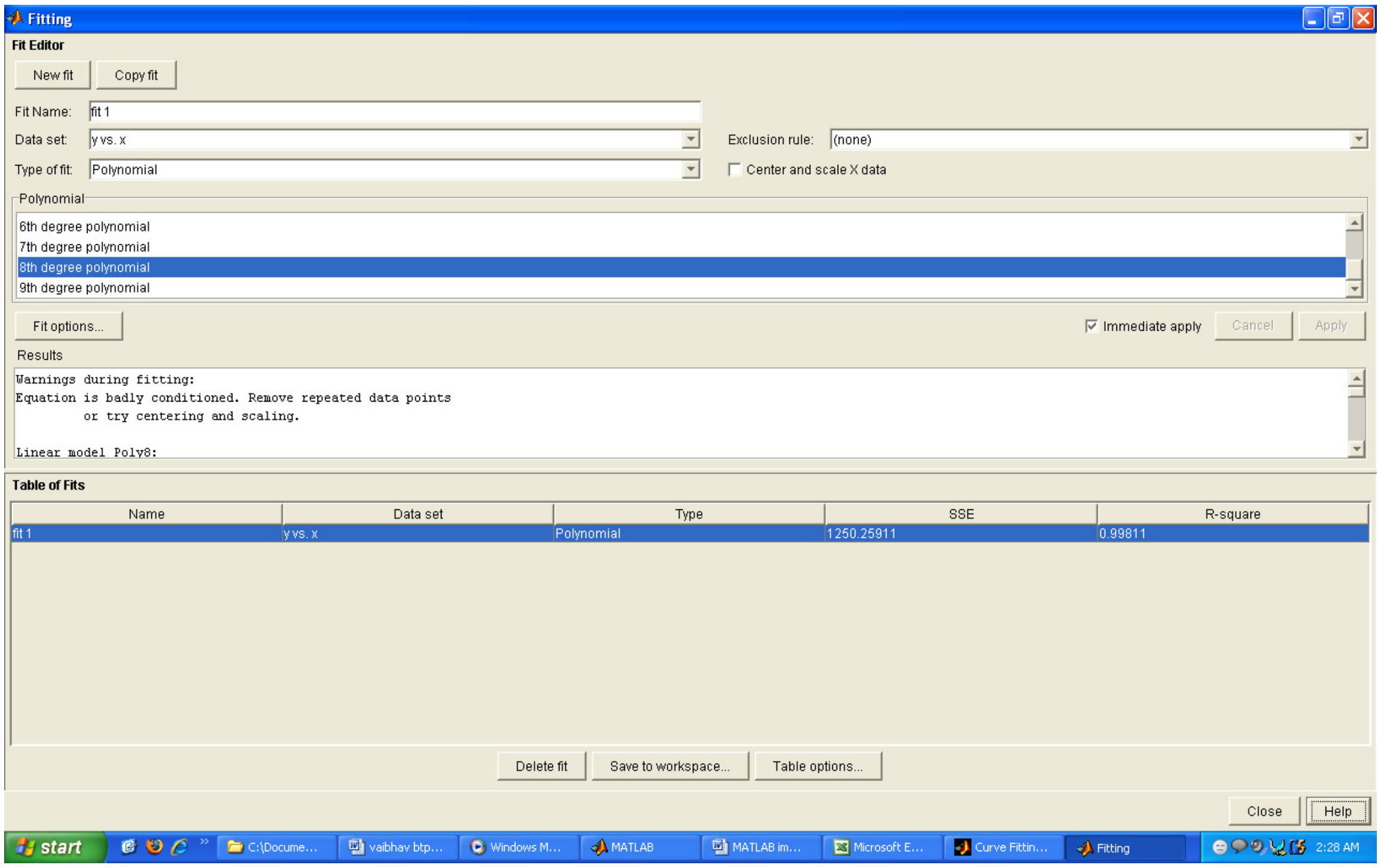

# Polynomial Approximation with an 8th degree polynomial

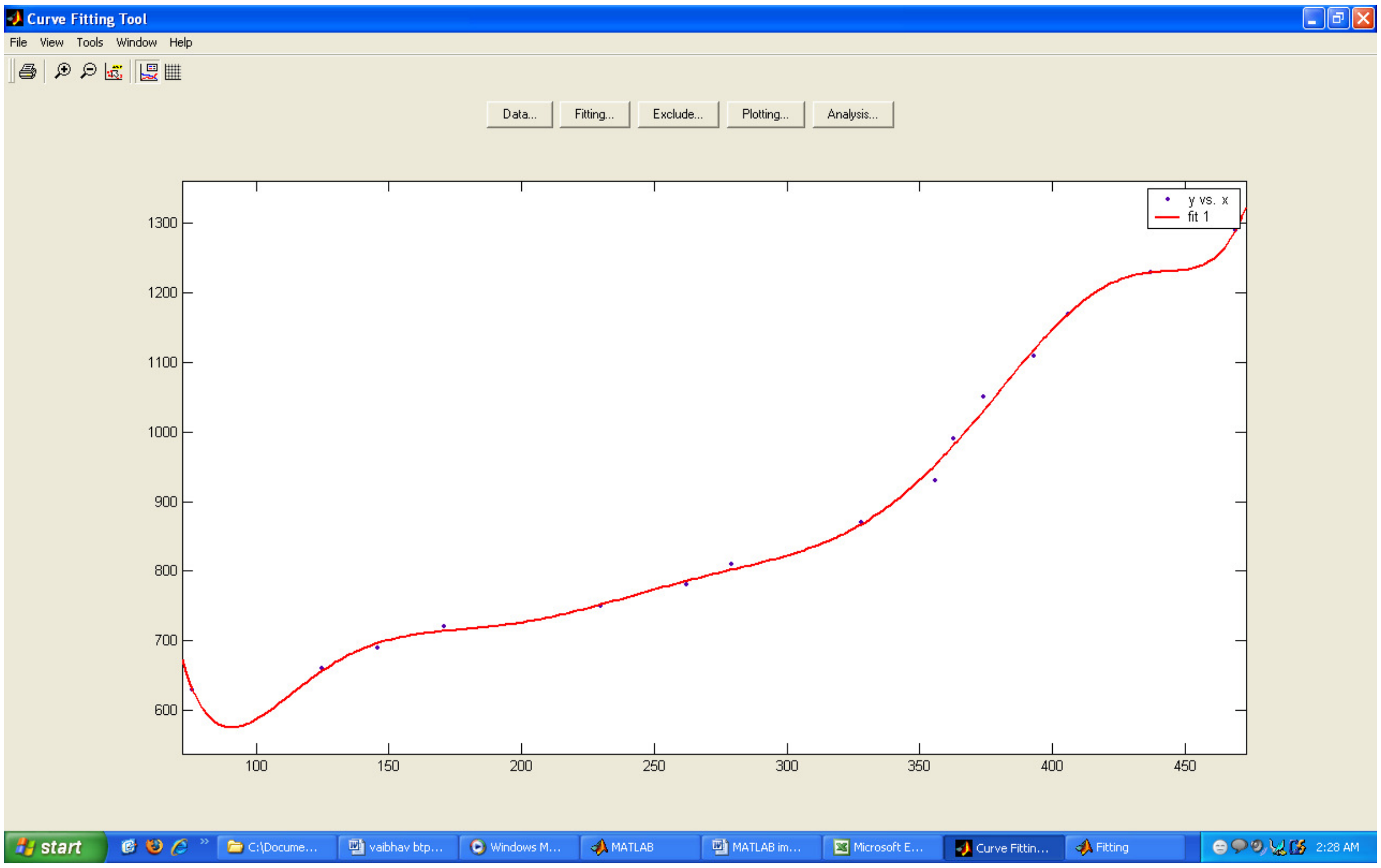

**Linear model:**

f(x) = p1*x^8 + p2*x^7 + p3*x^6 + p4*x^5 + p5*x^4 + p6*x^3 + p7*x^2 + p8*x + p9

**Coefficients:**

p1 = -1.231e-015 (-3.759e-015, 1.296e-015)
p2 = 3.828e-012 (-4.387e-012, 1.204e-011)
p3 = -5.13e-009 (-1.667e-008, 6.407e-009)
p4 = 3.87e-006 (-5.271e-006, 1.301e-005)
p5 = -0.001797 (-0.006264, 0.002671)
p6 = 0.5256 (-0.8529, 1.904)
p7 = -94.58 (-356.8, 167.6)
p8 = 9575 (-1.852e+004, 3.767e+004)
p9 = -4.173e+005 (-1.715e+006, 8.808e+005)

**Goodness of fit:**

RMSE: 13.68